# ON HEALTHCARE ROBOTS

Concepts, definitions, and considerations for healthcare robot governance

Eduard Fosch-Villaronga & Hadassah Drukarch

## Author details

**Dr. Eduard Fosch-Villaronga**, Assistant Professor, eLaw Center for Law and Digital Technologies, Leiden University, the Netherlands, e.fosch.villaronga@law.leidenuniv.nl.

**Hadassah Drukarch**, Research Assistant, eLaw Center for Law and Digital Technologies, Leiden University, the Netherlands, h.g.drukarch@law.leidenuniv.nl.

**May 2021**

# Disclaimer

This report was funded by the LEaDing Fellows Marie Curie COFUND fellowship, a project that has received funding from the European Union's Horizon 2020 research and innovation programme under the Marie Skłodowska-Curie Grant Agreement No. 707404.

This report's contents are the authors' sole responsibility and do not necessarily reflect the views of the European Commission. The opinions expressed here are those of the authors only and do not necessarily reflect the authors' organizations.

Contents of this publication may be quoted or reproduced, provided that the source of information is acknowledged.

# Acknowledgments

This report was funded by the LEaDing Fellows Marie Curie COFUND fellowship, a project that has received funding from the European Union's Horizon 2020 research and innovation programme under the Marie Skłodowska-Curie Grant Agreement No. 707404.

The authors would like to thank Mrs. Aline-Priscilla Messi and Mrs. Rosalie ten Wolde for their help on the first draft of some parts of this report during their student assistantship as part of the Honors College Law at Leiden University during March-Aug 2020. The authors would also like to thank T.J.C. Faes, Associate Professor in Biomedical Engineering, Dept. Radiology and Nuclear Medicine, Amsterdam UMC, for his valuable feedback on an earlier draft.

# Abstract

**The rise of healthcare robotics**
Robotics have increased productivity and resource efficiency in the industrial and retail sectors, and now there is an emerging interest in realizing a comparable transformation in healthcare. Robotics and artificial intelligence (AI) are some of the latest promising technologies expected to increase the quality and safety of care while simultaneously restraining expenditure and, recently, reducing human contact too. Healthcare robots are likely to be deployed at an unprecedented rate due to their reduced cost and increasing capabilities such as carrying out medical interventions, supporting biomedical research and clinical practice, conducting therapy with children, or keeping the elderly company.

**The lack of healthcare robot policy**
Although healthcare is a remarkably sensitive domain of application, and systems that exert direct control over the world can cause harm in a way that humans cannot necessarily correct or oversee, it is still unclear whether and how healthcare robots are currently regulated or should be regulated. Existing regulations are primarily unprepared to provide guidance for such a rapidly evolving field and accommodate devices that rely on machine learning and AI. Moreover, the field of healthcare robotics is very rich and extensive, but it is still very much scattered and unclear in terms of definitions, medical and technical classifications, product characteristics, purpose, and intended use. As a result, these devices often navigate between the medical device regulation or other non-medical norms, such as the ISO personal care standard. Before regulating the field of healthcare robots, it is therefore essential to map the major state-of-the-art developments in healthcare robotics, their capabilities and applications, and the challenges we face as a result of their integration within the healthcare environment.

**Our contribution to the policymaking debate on healthcare robots and AI technologies**
This contribution fills in this gap and lack of clarity currently experienced within healthcare robotics and its governance by providing a structured overview of and further elaboration on the main categories now established, their intended purpose, use, and main characteristics. We explicitly focus on surgical, assistive, and service robots to rightfully match the definition of healthcare as the organized provision of medical care to individuals, including efforts to maintain, treat, or restore physical, mental, or emotional well-being. We complement these findings with policy recommendations to help policymakers unravel an optimal regulatory framing for healthcare robot technologies.

# Abbreviations

| | |
|---|---|
| **ADL** | Activities of the Daily Living |
| **AG** | Advocacy Groups |
| **AGV** | Automated Guided Vehicles |
| **AI** | Artificial Intelligence |
| **ANN** | Artificial Neural Networks |
| **BCI** | Brain-Computer Interface |
| **CG** | Caregivers |
| **CL** | Clinicians |
| **DRU** | Direct Robot Users |
| **FDA** | Food and Drug Administration |
| **FRS** | Fundamentals of Robotic Surgery |
| **ESW** | Environmental Service Workers |
| **GDPR** | General Data Protection Regulation |
| **GPS** | Global Positioning Systems |
| **HA** | Health Administrators |
| **HAI** | Healthcare-Associated Infections |
| **HMI** | Human-Machine Interface |
| **HNR** | Humanoid Nursing Robots |
| **HRI** | Human-Robot Interaction |
| **HSR** | Healthcare Service Robot |
| **IC** | Insurers |
| **IFR** | International Federation of Robotics |
| **ISO** | International Standardisation Organisation |
| **MDR** | Medical Device Regulation |
| **MIS** | Minimally Invasive Surgery |
| **MSR** | Mobile Servant Robots |
| **OECD** | Organisation for Economic Co-operation and Development |
| **PAR** | Physically Assistive Robot |
| **PCN** | Physically, cognitively, or neurologically |
| **PM** | Policy Makers |
| **PPE** | Personal Protective Equipment |
| **RAS** | Robotic-Assisted Surgery |
| **RASE** | Robotically Assisted Surgical Equipment |
| **RIC** | Remote Inpatient Care |
| **RM** | Robot Makers |
| **ROC** | Remote Outpatient Care |
| **SAE** | Society of Automotive Engineers |
| **SAR** | Socially Assistive Robots |
| **SCI** | Spinal Cord Injury |
| **SSI** | Surgical Site Infection |
| **UV** | Ultraviolet |
| **VR** | Virtual Reality |
| **WHO** | World Health Organisation |

# Contents

# The "Healthcare Robots and AI" project and report

## The project "Healthcare Robots and AI"

Healthcare robots and AI is a project that investigates the legal and regulatory aspects of healthcare robot and Artificial Intelligence (AI) technologies.

The integration of physical robotic systems with cloud-based services in healthcare settings is accelerating. This research project investigates the legal and regulatory implications of the growing interdependence and interactions of tangible and virtual elements in cyber-physical systems for healthcare purposes. Typical examples of such cyber-physical systems are cognitive therapeutic robots, physical rehabilitation robots, assistive and surgery robots. As these technological developments may raise different types of issues, ranging from privacy invasion to autonomy suppression or human-human interaction decrease, there may be a need for (some forms of) regulation. This project highlights specific problems and challenges in regulating complex and dynamic cyber-physical ecosystems in concrete healthcare applications and explores potential solutions.

This project contributes to existing research in this area by assessing how recent legal and regulatory initiatives governing robots apply to healthcare robots and artificial intelligence technologies. The project explores rehabilitation robots (physical and cognitive), assistive, and surgery robots, which are cyber-physical systems that involve multiple parties and the use of various physical devices (with different embodiments) and virtual services to illustrate practical challenges. The project also considers novel applications of sexual robots in healthcare. In this way, the project seeks to provide a thoughtful, innovative, and thorough review of legal and regulatory issues concerning cyber-physical systems in healthcare settings. This investigation is in line with previous research (Fosch Villaronga, 2019) on 'Healthcare, Robots, and the Law. Regulating Automation in Personal Care' with the hope that this will inform the policy debate and set the scene for further research in the field of law and healthcare robots.

This project is part of the LEaDing Postdocs Fellowship Programme, a project funded by the European Union's Horizon 2020 research and innovation program under the Marie Skłodowska-Curie grant agreement No 707404. Part of this fellowship is carried out in collaboration with AiTech Center at TU Delft, which is a multidisciplinary research program on awareness, concepts, and design & engineering of autonomous technology under meaningful human control. The output of this research can be found here.

## Goals of this report

The field of healthcare robotics is very rich and extensive. However, it is still very much scattered and unclear in terms of definitions, medical and technical classifications, product characteristics, purpose, and intended use. Unclear definitions and categories adversely impact the understanding of how the legislation applies to concrete robot applications and how it may hamper compliance processes and safety. Moreover, the healthcare robotics field continuously evolves, challenging policymakers' job of framing this revolution to ensure user rights and safety.

Through this contribution, we attempt to fill in some of the gaps and lack of clarity currently experienced within the field of healthcare robotics by providing a structured overview of and further elaboration on the main categories now established, their intended purpose, use, and main characteristics. This could complement a recent publication from the Joint Research Center Science for Policy Report on Artificial Intelligence in Medicine and Healthcare: applications, availability and societal impact (Gómez-González, & Gómez, 2020), because we focus on surgical, assistive, and service robots to rightfully match the definition of healthcare as the organized provision of medical care to individuals, including efforts to maintain, treat, or restore physical, mental, or emotional well-being. Emerging healthcare robots were taken into account for parts of this research, especially within the field of socially and physically assistive robots.

## A list of policy recommendations

Our findings provide the following policy recommendations for healthcare robot governance:

1. Healthcare robots differ from other types of robots and deserve special regulatory attention.

2. Legal frameworks should acknowledge the high-risk nature of robotic systems for healthcare purposes.

3. Clarity on the medical device categorization for healthcare robot types, including surgical, physically and socially assistive, and healthcare service robots is needed at risk that they will be repurposed without the appropriate safeguards.

4. Universal standards for progressive autonomy levels for healthcare robots should be defined to increase legal certainty with respect to responsibility allocation in highly complex human-robot interaction contexts, also in the medical field.

5. The parameters in which the need for healthcare robots is assessed cannot be resource efficiency and increased productivity as with industrial robots only, other aspects as the implications for society need to be considered.

6. To address safety comprehensively, healthcare robots demand a broader understanding of safety, extending beyond physical interaction, but covering aspects such as cybersecurity, temporal aspects, societal dimensions and mental health.

7. Embodied healthcare robots can also exacerbate existing biases against certain groups and, therefore, their design, implementation, and use should account for diversity and inclusion.

# On healthcare robots and their governance

Robotics have increased productivity and resource efficiency in the industrial and retail sectors, and now there is an emerging interest in realizing a comparable transformation in healthcare (Cresswell, Cunningham-Burley & Sheikh, 2018). Robotics and AI are some of the latest promising technologies expected to increase the quality and safety of care while simultaneously restraining expenditure, given their success in the industrial sector (Riek, 2017; Cresswell, Cunningham-Burley & Sheikh, 2018). In this respect, healthcare robots are likely to be deployed to this end at an unprecedented rate (Simshaw et al., 2015) as a result of their reduced cost and their increased roles and capacities (COMEST, 2017) to perform medical interventions (Nouaille et al., 2017), support impaired patients (Tucker et al., 2015), provide therapy to children (Scassellati, Admoni & Matarić, 2012) or keep the elderly company (Broekens, Heerink & Rosendal, 2009).

Despite several clear benefits, systems that exert direct control over the world can cause harm in a way that humans cannot necessarily correct or oversee (Amodei et al., 2016). For instance, safety issues such as injury or death arise if robot surgeons power down mid-operation or operate unintendedly (Alemzadeh et al., 2016; Ferrarese, 2016). Security vulnerabilities allow unauthorized users to remotely access and control robots, potentially harming patients (FDA, 2020), blurring the discernment of who is responsible for those cases. Although therapeutic robots are not a standardized form of care, their continuous use could increase isolation, deception, and loss of dignity (Sharkey & Sharkey, 2012; Sharkey, 2014; Zardiashvili & Fosch-Villaronga, 2020). Moreover, while the collection of sensitive information is critical for the development of robots and to support patients adequately, collecting such amounts of data may infringe upon a person's right to privacy and autonomy (Syrdal et al., 2007; Bertolini, & Aiello, 2018; Ienca & Fosch-Villaronga, 2019). Beyond this, robots are cyber-physical entities, with self-learning capabilities and adaptive behavior, and automated decision-making processes that are not always transparent (Poulsen, Burmeister & Tien, 2018; Fosch-Villaronga et al., 2018). Robots also interact with vulnerable users in many ways, including cognitively vulnerable persons. They may even offer unimaginable possibilities, e.g., helping realize the sexual rights of disabled people (Di Nucci, 2017; Fosch-Villaronga & Poulsen, 2020). This all pushes the boundaries of our current understanding of how these robots entail risks for users and, therefore, how they should be regulated.

Although healthcare is a strikingly sensitive domain of application, it is still unclear whether and how healthcare robots are currently regulated or should be regulated (Simshaw et al., 2015; Leenes et al., 2017; Fosch-Villaronga, 2019). Some argue that robot-oriented regulations seem premature, misguided, or even dangerous because robots are at an early stage, and a misconceived regulation thereof could hinder the development of such technology, thereby preventing their significant societal benefits (Brundage, & Bryson, 2016). For instance, the belief that robots could dehumanize caring practices (European Parliament 2017, 2019) might prevent developing feeding-robots that allow for increased privacy during mealtime (Herlant, 2018); or robots for the blind, which could improve users' autonomy and help assistance-dogs avoid welfare-threatening punishment-based training (Bremhorst et al., 2018). Still, existing regulations are mostly unprepared to accommodate devices that rely on machine learning and AI. These devices may modify their functioning during their lifecycle, but current legislation ensures safety risks present when the device enters the market. In this respect, the law needs to be clearer on handling the progressive and adaptive AI outcomes that may be utterly different subsequent market entrance (European Commission, 2020a; FDA, 2019).

An example of how unprepared legislation is to govern this field can be found in the proposed AI Act 2021. In April 2021, the European institutions released a proposal for a regulation laying down harmonized rules on artificial intelligence (AI Act, 2021). Before, there was an absence of specific AI or robot regulation in which clear procedures, boundaries, and requirements for AI developers are explained, challenging how they can integrate these considerations into their design to make them safe (Holder et al., 2016; Fosch-Villaronga, 2019). The AI Act (2021) establishes as 'high-risk' those 'AI systems that pose significant risks to the health and safety or fundamental rights of persons' (p. 3). As such, one would think that algorithmic systems that generate health-related outcomes will generally be considered high-risk. However, while the AI Act (2021) in Annex III lists high-risk applications, they do not include any application considering healthcare or medicine.[1] Not being categorized as 'high-risk' means that the requirements that would typically apply to high-risk systems do not apply to such applications. These requirements refer to the high quality data, documentation and traceability, transparency, human oversight, accuracy and robustness, which are strictly necessary to mitigate the risks to fundamental rights and safety posed by AI (AI Act, 2021). Moreover, it also means that non high-risk AI systems will not have to comply with a set of horizontal mandatory requirements for trustworthy AI that the High-level Expert Group on AI established in 2019. In these guidelines, they proposed a 'Trustworthy AI assessment list' aimed at operationalizing the critical requirements of (1) human agency and oversight, (2) technical robustness and safety, (3) privacy and data governance, (4) transparency, (5) diversity, non-discrimination, and fairness, (6) environmental and societal well-being, and (7) accountability (HLEG AI, 2019).

Before regulating the field of healthcare robots, it is essential to map the major state of the art developments in healthcare robotics, their capabilities and applications, and the challenges we face as a result of their integration within the healthcare environment. Our contribution seeks to bring comprehension to the field of healthcare robotics by providing a clear definition and classification of healthcare robot types, contexts of use, and considerations to assist in the development of and to help unravel a subsequent optimal regulatory framing. Our contribution thereby complements existing classification efforts from the International Federation of Robots (IFR), which, in a joint effort started in 1995 together with the United Nations Economic Commission for Europe (UNECE), engaged in working out a first service robot definition and classification scheme. This definition and classification scheme has been absorbed by the current ISO Technical Committee 184/Subcommittee 2 and resulted in a novel ISO-Standard 8373, which became effective in 2012. ISO 8373:2012 specifies vocabulary used in relation to robots and robotic devices operating in industrial and non-industrial environments, including healthcare (Heagele et al., 2016). Healthcare is the organized provision of medical care to individuals, and it includes efforts to maintain, treat, or restore physical, mental or emotional well-being. That is why, in this contribution, we focus on physically and socially assistive robots, surgical robots, and healthcare service robots. As advancements in robotics do not occur in every healthcare area, we acknowledge that our contribution may appear to provide a fragmented vision of healthcare. However, healthcare robotics is a continually evolving field that contains medical device misclassifications and misleading labels that confuse and challenge the safety of the provision of care by robots. While healthcare robots may call for new ways to understand the adequacy of existing norms in different areas and moments, our effort is a stepping stone toward understanding the policy implications that existing and new types and contexts of healthcare robots (may) have.

---

[1] The high-risk categories, according to annex III AI Act 2021 are: 1) Biometric identification and categorisation of natural persons, 2) Management and operation of critical infrastructure, 3) Education and vocational training, 4) Employment, workers management and access to self-employment 5) Access to and enjoyment of essential private services and public services and benefits, 6) Law enforcement, 7) Migration, asylum and border control management, 8) Administration of justice and democratic processes. *See* https://ec.europa.eu/newsroom/dae/document.cfm?doc_id=75789.

## Methods

For the purpose of this contribution, we conducted a desktop and literature review and compiled sources for six months, spanning from March to July 2020. These sources include research articles, web pages, and product catalogs, retrieved from online academic databases and web search results on Google and Google Scholar. For the research conducted on and analysis of surgical robots, we considered only commercially available robot applications. The field of surgical robots is already sufficiently developed to obtain a clear overview of its current state of the art and the direction in which the area is developing, without this necessarily being a limiting factor to the research conducted and the analysis based thereon.

For the study conducted on and analysis of assistive robots, both academically and commercially available robot applications were considered as the field of assistive robotics is very much still in the beginning stages of its development. Finding robots that corresponded to the definition we attributed to *socially assistive robots* was initially difficult due to the terminological "free for all" used within different social robotics branches. By including robots developed for specific (academically related) projects, we aimed to define the field of assistive robotics better and obtain an overview of the direction in which the area is developing, as narrowing robots down to commercially available applications proved to be too limiting.

We considered academic and commercially available robot applications for the research on and analysis of healthcare service robots. While the field of healthcare service robots is already commercially developed to a significant degree, external developments (e.g., the outbreak of the COVID-19 pandemic) have given rise to research on and the development of new or improved service robots in the field of healthcare. Given the broad definition of healthcare service robots, a wide variety of healthcare service robots exist. The literature pertaining to healthcare service robots lacks any specific categorization of healthcare service robots. Therefore, a description of the relevant robots is usually provided through their characteristics, which also proved to be a more practical search strategy to access specific information about their features, purpose, context, and (intended) use.

For this research, we constructed preliminary research tables based on the retrieved information from a set of primary keywords for each of the considered robot types - surgical, assistive, and Healthcare Service Robots. The tables were further refined by using a set of secondary keywords, which formed the basis for a more extensive, specific, and in-depth analysis of types, contexts of use, and characteristics (*see* table 1). For this research, we thus applied a two-step keyword literature review based on which preliminary and secondary tables were constructed. From this, we derived our findings, and on the basis thereof, we formulated our conclusions as presented in this report (*see* figure 1).

| Healthcare robot category | Primary keywords (examples) | Secondary keywords (examples) |
|---|---|---|

| | | | |
|---|---|---|---|
| **Surgical robots** | | surgery robots/surgical robots; robotic surgery; surgery automation; autonomy levels robotic surgery; minimally invasive surgery; telesurgery; telesurgical robots; automated surgery; automated surgical robots; supervisory-controlled surgery; supervisory-controlled surgical robots; shared-controlled surgery; shared-controlled surgical robots | da Vinci System; CyberKnife System; ROBODOC; ZEUS Robotic Surgical System; Senhance Surgical System; Mako Smart Robotics; Microsure; Flex Robotic System; |
| **Assistive robots** | **Socially Assistive Robots** | socially assistive robots; assistive robots; and therapy robots; robot-assisted therapy; therapeutic robots; care robots; autism therapy robots; social robots for therapy; conversational robots | Huggable robot; Bandit robot; Pleo robot; Paro; AIBO; Pepper; Kaspar; Probo; Milo; Huggable; Kiwi; Kompaï; |
| | **Physically Assistive Robots** | assistive robots; robotherapy; therapeutic robots; physically assistive robots; rehabilitation robots; active-assisted robot therapy; robot nurses lifting aid; lifting robots; smart/intelligent/power/robotic wheelchairs; robot nurses; robot nursing assistant; | Mealtime Partner Assistive Dining Device; My Spoon; iArm; iEAT Robot ; iFLOAT Arm Support; The KINOVA® Assistive Technologies Series; EDAN; OBI® eating device; eLEGS; ROBERT®; Baxter Research Robot; EksoNR; Neater Eater; Andago®; REX; ReWalk Personal 6.0; |
| **Healthcare Service Robots** | | service robots healthcare/healthcare service robots; medical service robots; routine task robots; nurse assisting robots; robot nurses; delivery robots; disinfectant robots; telepresence robots; routine task robots; | Relay; HOSPI; Giraff; Roomba i7; Care-o-bot; corona robot; covid-19 healthcare robot; disinfectant robots covid; |

**Table 1.** Primary and secondary research keywords

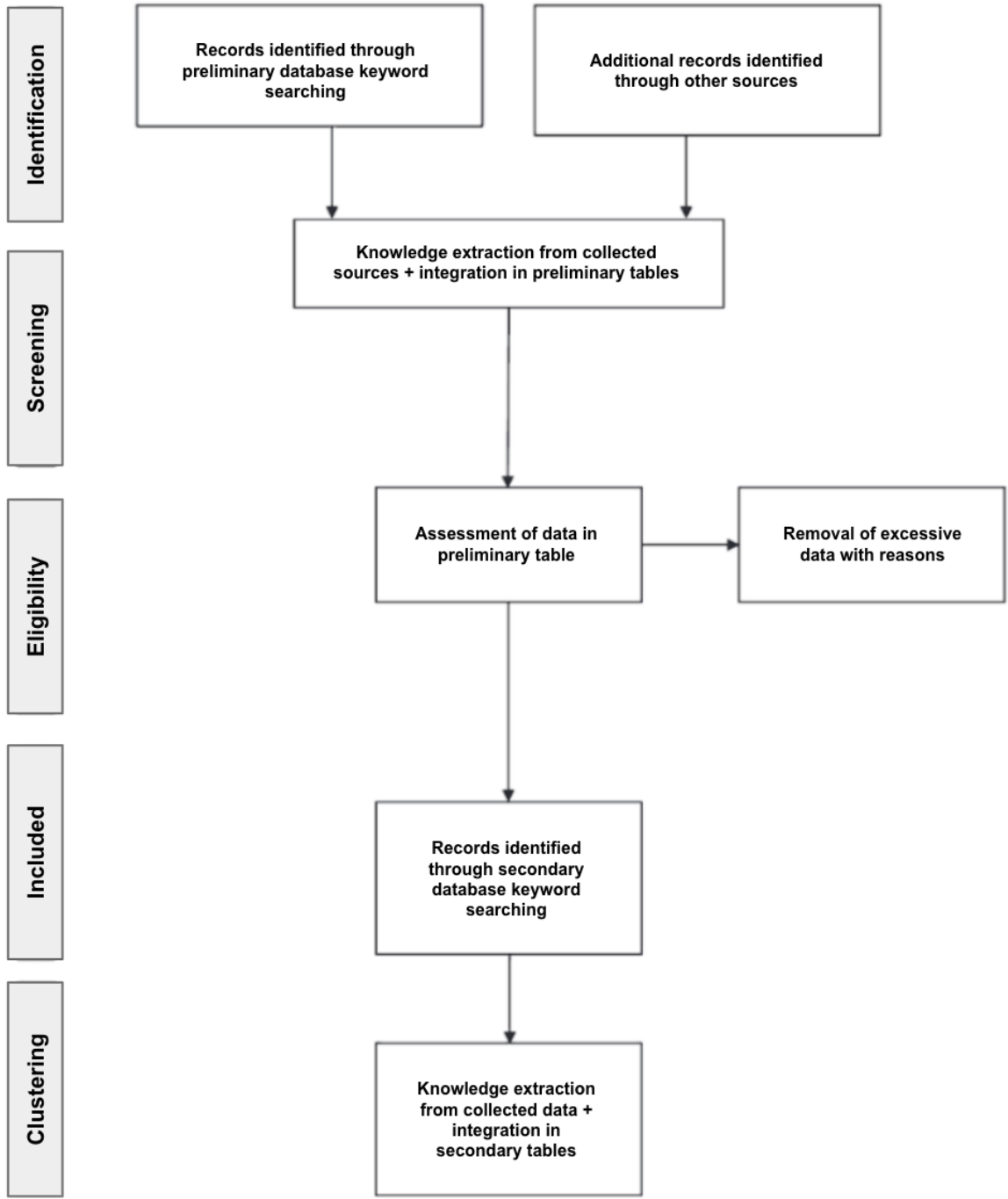

**Figure 1.** Literature review flow diagram

Thus, the preliminary tables formed the basis for the secondary research tables, which provide a more extensive and structured overview of several examples of widely recognized robots used within the field of healthcare, more specifically the categories of healthcare robots we take into consideration. These tables include various elements, including robot names, manufacturing companies, purposes, contexts of use, robot characteristics, release dates, and control mechanisms. The findings we extracted from these tables and the conclusions we came to based thereon form the basis of our attempt to map the major state of the art developments in healthcare robotics, their capabilities, and applications, as well as the challenges we face as a result of their integration within the healthcare environment.

# Chapter 1: Defining healthcare robotics and their categories

1.1. The importance of definitions

Precise terminology has always been important (Chapin, 1939; Scott, 2001). Even if definitions are not an outcome but a single step in a long process of understanding, the terminology is still the primary reference we acknowledge to define concepts, ideas, or notions (Fetzer, Shatz, Schlesinger, 2012). In general terms, therefore, definitions serve to create clarity and avoid misunderstandings when discussing a particular subject. However, not all concepts are easy to describe. For instance, everyone knows what *emotion* is until asked to give a definition (Fehr and Russell, 1984). As experts are asked to define it, there are also many definitions of *intelligence* (Gregory and Zangwill, 1987). Something similar happens with the word *robot* or with the word *artificial intelligence* (Fosch-Villaronga, 2019).

Especially within the field of law, definitions play a pivotal role, functioning as a mechanism for avoiding ambiguity in interpretation, and most importantly, warranting the application of the law to a case (Macagno, 2010). As far as applying the law is concerned, the elaboration of definitions in the legal domain has a dual nature. In the first place, definitions support the classification of entities in legal categories and consequently warrant the enforcement of legal consequences, especially in fields that have not yet extensively been the subject of legal research. Therefore, all-encompassing legal definitions for healthcare robots are yet to be established (Simshaw et al., 2015). Secondly, definitions in the legal domain can, in addition to a delimitation and a clarification into legal categories, broaden certain regulations' scope. When a broad definition of a concept is formulated, it can apply to a wide range of situations, including unforeseen circumstances. Law and regulations are generally fairly rigid. Given the rapid development in today's society, it is not always possible to foresee every situation in which specific regulations should apply. By formulating broad definitions, a solution can be found in these kinds of cases, and as a result, the applicability of the regulations in question is not at issue. However, not all concepts are easy to describe, especially when law and technology interact.

Whether or not a definition is required essentially depends on the level of legal certainty a concept would provide (Fosch-Villaronga & Millard, 2019). The law does not necessarily need to define a thing to regulate it. For instance, the concept of *electricity* is not defined in the Directive 2009/72/EC of the European Parliament and of the Council of 13 July 2009 concerning common rules for electricity's internal market. However, if robot technology is to be regulated definitively and explicitly, a clear definition is likely to increase legal clarity and certainty. Legislators might find that defining a robot would most likely be the preferable course of action, but even more, so that it is necessary to have such a definition in place. Nevertheless, the legal and ethical communities have no generally accepted definition for the term *robot*. For this contribution, we define a *robot* as 'a movable machine that performs tasks either automatically or with a degree of autonomy' (ISO 8373:2012; Richards, & Smart, 2016; Fosch-Villaronga & Millard, 2019).

## 1.2. Healthcare robot categories and definitions

Nowadays, robots have not merely become routine in the world of manufacturing and other repetitive labor tasks. They have also penetrated other fields, including healthcare, where they are used for entirely different environments and tasks, involving direct interaction with human users, in the surgical theater, the rehabilitation center, and the family room (Okamura, Mataric, & Christensen, 2008). Research points out that an estimated 20% of the world's population experiences difficulties with physical, cognitive, or sensory functioning, mental health, or behavioral health - either temporary or permanent, acute or chronic, and subject to change throughout one's lifespan (Riek, 2016). A significant number of these individuals experience severe difficulties with Activities of the Daily Living (ADL) (Riek, 2017). Recent developments in healthcare robotics have fundamentally changed how the medical and healthcare environment's functioning is perceived. The societal drivers for improved health care that can be addressed by robotic technology, broadly, lie in two categories: the wish to broaden access to healthcare and to improve prevention and patient outcomes (Cresswell, Cunningham-Burley & Sheikh, 2018). Technological advances in robotics have shown to have clear potential for stimulating the development of new medical treatments for a wide variety of diseases and disorders, for improving both the standard and accessibility of care, for enhancing patient health outcomes, and for filling quantitative care gaps, supporting caregivers, and aiding healthcare workers (Kim, Gu, & Heo, 2016). The spectrum of robotic system niches in medicine and healthcare currently spans a wide range of environments, user populations, and interaction modalities.

The European Foresight Monitoring Network (EFMN) (2008) defined *healthcare robots* as 'systems able to perform coordinated mechatronic actions (force or movement exertions) based on processing information acquired through sensor technology, to support the functioning of impaired individuals, medical interventions, care and rehabilitation of patients and also individuals in prevention programs.' Robots used within the field of healthcare encompass varying degrees of autonomy and broadly include affiliated technology, including sensor systems, algorithms for processing data, and cloud services (Riek, 2016; Fosch-Villaronga & Millard, 2019).

Over time, the Policy Department for Economic, Scientific, and Quality of Life Policies of the European Parliament identified 'the most interesting applications of healthcare robots', which include robotic surgery, care, and socially assistive robots, rehabilitation systems, and training for healthcare workers (Dolic, Castro, & Moarcas, 2019).

Analysis of past and current performance of robotics within the field of healthcare have proved that there is an incredible opportunity for robotics technology to help fill care gaps, to aid healthcare workers, for use in physical and cognitive rehabilitation, surgery, telemedicine, drug delivery, and patient management (Riek, 2017). In this contribution, we, therefore, focus on three main categories of healthcare robots: surgical robots, assistive robots, and Healthcare Service Robots (HSR) (see Table 2). We distinguish between Physically Assistive Robots (PAR) and Socially Assistive Robots (SAR) within the context of assistive robots.

| Healthcare robot categories | | Definition |
|---|---|---|
| **Surgical robots** | | Service robots supporting surgeons during surgical procedures. |
| **Assistive robots** | **Socially Assistive Robots** | Service robots assisting users through social interaction. |
| | **Physically Assistive Robots** | Service robots supporting users through physical interaction. |
| **Healthcare Service Robots** | | Service robots in a healthcare setting performing tasks useful to the facility and the medical staff. |

**Table 2.** Healthcare robot categories and definitions.

# Chapter 2: Defining the healthcare robot ecosystem

The healthcare ecosystem is the network of stakeholders, processes, and materials necessary for the treatment of an ailment by way of medical intervention on a patient (de Vries 2016). The extensive list of robotics stakeholders in general used in society identified in the European project 'RoboLaw', which includes producers and employers of robots, insurance companies, trade-unions, user associations, professional users, and policymakers (Palmerini et al., 2014), to a certain extent, can also be identified in the field of healthcare robots (Fosch-Villaronga, et al., 2021), although this field calls for a more specific approach because of the many parties involved and the healthcare setting's particular nature.

Within the field of healthcare robots, several stakeholders can be identified (see Table 3). Many different actors use healthcare robots within a healthcare setting: doctors, medical professionals, patients, family members, caregivers, healthcare providers, or even technology providers. All these stakeholders have similar goals, although they experience healthcare from different viewpoints: providing (medical) care and independence, preserving patients' dignity, and empowering those with special needs (Simshaw et al. 2016). A common and practical approach is to divide the stakeholders in healthcare robotics into primary, secondary, and tertiary stakeholders (Riek, 2017).

| Main category stakeholder | Subcategory stakeholder | Description |
|---|---|---|
| **Primary stakeholders** | Direct Robot Users (DRU) | Primary stakeholders use healthcare robotics on a regular or even daily basis. |
| | Clinicians (CL) | |
| | Caregivers (CG) | |
| **Secondary stakeholders** | Robot Makers (RM) | Secondary stakeholders are involved in using healthcare robotics, but will not directly use them themselves. |
| | Environmental Service Workers (ESW) | |
| | Health Administrators (HA) | |
| **Tertiary stakeholders** | Policy Makers (PM) | Tertiary stakeholders are those parties who have an interest in the use and deployment of healthcare robotics in society, although it is unlikely that they will use them directly. |
| | Insurers (IC) | |
| | Advocacy Groups (AG) | |

**Table 3.** Stakeholders in the field of healthcare robotics (Riek, 2017)

Within the primary stakeholders, Riek (2017) identifies Direct Robot Users (DRU), Clinicians (CL), and Caregivers (CG). The primary stakeholders are those who will use healthcare robotics on a regular or even daily basis. Within the category of secondary stakeholders, she identifies the Robot Makers (RM), the Environmental Service Workers (ESW), and the Health Administrators (HA). These secondary stakeholders are involved in using healthcare robotics but will not directly use them themselves. Finally, within the category of tertiary stakeholders, Riek (2017) distinguishes between Policy Makers (PM), Insurers (IC), and Advocacy Groups (AG). The tertiary stakeholders are those who are interested in the use and deployment of healthcare robotics in society, although it is unlikely that they will use them directly.

# Chapter 3: Healthcare Robot Categories

1. Surgical robots

Surgical robots are service robots supporting surgeons during surgical procedures. Since the mid-1980s, when the first robotic-assisted surgical procedures took place, surgical robotics has evolved into a highly dynamic and rapidly growing field of application and research, enjoying increasing clinical attention worldwide (Faust, 2007; Bergeles, & Yang, 2013; Lane, 2018). Initially introduced for a limited type of surgical procedures, nowadays, advances in ergonomics, computing power, hardware dexterity, safety, and ease of surgery allow for the rapid adoption and dissemination of new technologies for robotic-assisted surgical procedures. These include, in particular, an increasing amount of minimally invasive surgical operations, i.e., for operations that involve the insertion of a narrow laparoscopic device into the human body instead of having to open up the patient to that end (Sridhar et al., 2017). The current benefits of robotic surgery, or Robotic-assisted Surgery (RAS) - including, among other things, increased accuracy, precision, dexterity, tremor corrections, scaled motion, and haptic corrective feedback - result in less damage to the patient's body, more successful surgeries, and less invasive procedures that grant shorter patient recovery time and hospital stays, less pain, blood loss, noticeable scars and discomfort, and less risk of complications following the procedure (Boyraz et al., 2019; Jaffray, 2005). Still, the rapidly increasing demand resulting from the beneficial responses to their use and the consequent demonstration of their practical clinical potential leaves an extensive amount of room for further development and innovation. Since robots are devoid of shortcomings such as fatigue or momentary lapses of attention, they can perform repeated and tedious surgeries, enabling at the same time, the performance of surgical procedures that were previously considered impossible (Fosch-Villaronga et al., 2021). For instance, RAS could also help 'optimize the production, distribution, and use of the health workforce and infrastructure; allocate system resources more efficiently; and streamline care pathways and supply chains' in low- and middle-income countries (Reddy et al., 2016).

While RAS benefits abound, introducing a robot to the doctor-to-patient relationship changes how surgeries are performed. RAS extends the abilities of the doctor, but it also presents new challenges. A revision of 14 years of data from the FDA shows that robot surgeons can cause injury or death if they spontaneously power down mid-operation due to system errors or imaging problems (Alemzadeh et al., 2016). Broken or burnt robot pieces can fall into the patient, electric sparks may burn human tissue, and instruments may operate unintendedly, all of which may cause harm, including death (Alemzadeh et al., 2016). Moreover, as surgical robots' perception, decision-making power, and capacity to perform a task autonomously will increase, the surgeon's duties and oversight over the surgical procedure will inevitably change.

Moreover, other issues relating to cybersecurity and privacy will become more significant (Yang et al., 2017). Security vulnerabilities may allow unauthorized users to remotely access, control, and issue commands to robots, potentially causing harm to patients (FDA, 2020). Despite its widespread adoption for Minimally Invasive Surgery (MIS), a non-negligible number of technical difficulties and complications are still experienced during surgical procedures performed by surgical robots. To prevent or, at least, reduce such preventable incidents in the future, advanced techniques in the design and operation of robotic surgical systems and enhanced mechanisms for adverse event reporting ought to be adopted (Alemzadeh et al., 2016).

The robotic surgery ecosystem is a smaller ecosystem within the complex healthcare robots' ecosystem, comprising the surgeon, the nurses, and other staff members, who help the doctor during the surgical procedure, and the patients as the direct robot users (Fosch-Villaronga et al., 2021). The hospital administration also plays a role in the robotic surgery ecosystem, as it is the one looking for reliable measurements of processes costs, quality, and efficiency. Within this context, the use and role of robots affect and influence other stakeholders. However, some of them, among which the surgeon and support staff, will remain integral to the surgical environments for many functions, such as selecting the process parameters or positioning the patient, which further stresses the essential role humans still have in robot-mediated surgeries (Fosch-Villaronga et al., 2021).

Contemporary literature is rich in providing examples and applications of surgical robots and the current and predicted developments within surgical robotics (Bergeles & Yang, 2013). Still, the field is very much scattered, and a clear, concise definition for 'surgical robots' is still lacking. In this section, we fill in the gaps and lack of clarity currently experienced in surgical robotics by providing a structured overview of and further elaboration on the main categories currently established within the field of surgical robots, their purpose, context of use, and main characteristics. For this purpose, we define *surgical robots* as 'service robots that support surgeons during surgical procedures (Boyraz et al., 2019), allowing for more accurate and minimally invasive interventions'.

## 1.1. Surgical robots ecosystem

### 1.1.1. Surgical robots' state of the art

Robots used in surgery are not always surgical robots (Chinzei, 2019). Medical devices within the definition of robots exist in current surgeries like robot-shaped actuated operating tables or robotized microscopes. However, these two are usually not considered surgical robots. A *robotic surgical instrument* is 'an invasive device with an applied part, intended to be manipulated by Robotically-Assisted Surgical Equipment (RASE) to perform surgery tasks' (IEC 80601-2-77:2019).

We distinguish between the traditional types of surgical procedures in categorizing surgical robots, namely open surgery and closed surgery. *Open surgery* is the traditional form of surgery, which primarily refers to the highly invasive procedure of making an - often large - incision and cutting the skin and human tissue so that the surgeon has a full view of the structures and organs involved. Based on his medical assessment, the surgeon can determine and perform the necessary surgical procedures. While open surgery is generally considered a safe and effective type of surgery, it causes longer hospital stays, longer recovery periods, larger scars, more pain, and higher risks of complications (e.g., bleeding and infections). On the contrary, *closed surgery* refers to the minimally invasive technique involved in surgery that allows surgeons to performs surgical procedures by providing them access to the patient's body either through the body's natural openings or through small incisions in the body, and is only suitable under particular conditions (e.g., when there is no particular urgency or when the human capabilities lack the necessary precision). During the last three decades, MIS has influenced the techniques used in almost the entire surgical medicine field. The new approaches resulting from this allow surgeons to use various techniques to operate with less damage to the patient's body than would be the case with open surgery. As a result, MIS is generally associated with less pain and discomfort, shorter hospital stays, quicker recovery times, smaller scars, and less risk of complications following the procedure (Jaffray, 2005).

The continuing developments in MIS have led to the replacement of conventional surgery with minimally invasive surgical procedures. Still, they have also prompted surgeons to reevaluate conventional approaches to surgery. However, the introduction of these new approaches has, in some respects, led to significant drawbacks, e.g., prolonged learning curves for surgeons, increasing costs due to the (high) investments needed to acquire the necessary equipment and instruments, and longer operating times (Fuchs, 2002). Nowadays, many surgical techniques fall within the scope of MIS. RAS typically falls within the scope of MIS (Boyraz et al., 2019), the difference here being that instead of the surgeon manually operating instruments, they - as primary users of the robotic surgical systems - are supported or replaced by the power and precision of high-tech robotic systems.

Robotic systems are beginning to equal human specialists at precision surgical tasks. They may even outperform human surgeons in precision, control, efficiency, and accuracy, although it is still some time away before this applies to all surgical procedures. Increasingly autonomous robotic assistance levels allow intricate surgical feats to be performed without the surgeon worrying that their hands might slip or their grip falter (Svoboda, 2019). However, in surgeries that are very high volume, human surgeons are still much better than robots at weighing their experience to make complex surgical judgments and develop contextual understanding, especially when faced with unexpected situations and circumstances (Svoboda, 2019). It is precisely this power and accuracy that increasingly allows robotic systems to perform MIS, a type of surgery which, by its very nature, requires a high level of precision. The following table provides some examples of surgical procedures currently performed with the help of a robot:

| Surgical procedures performed with a robot | |
| --- | --- |
| Cardiac surgery | Ocular surgery |
| Cosmetic surgery | Orthopedic surgery |
| Dental surgery | Otorhinolaryngology |
| Endocrine surgery | Plastic and reconstructive surgery |
| Endoscopic surgery | Thoracic surgery |
| Gastrointestinal surgery | Urology |
| Gynecology | Vascular surgery |

**Table 4**. Surgical procedures performed with a robot

The field of RAS is rich in development and innovation. Surgical robots are used in different medical areas, usually on a spectrum that ranges from surgical robots that are more generic in nature, such as the da Vinci System®, to highly specific surgical robots, such as the PRECEYES Surgical System (R2D2). Depending on their capability and level of autonomy, surgical robots may be used for surgical procedures, ranging from less complicated surgeries on rigid body parts to more complex surgeries on soft human tissue (Prabu, Narmadha & Jeyaprakash, 2014). The technology incorporated and the surgical robot's embodiment play an essential role in the performance of surgical procedure's. Extensive research on the current state of the art of surgical robots shows that surgical robots' main characteristics include robotic arm(s) used to mimic and extend human movement, cutting instruments, cameras, and X-ray systems. In addition, they comprise surgeon consoles and probes, and mobile compartments and tools. In practice, robotic platforms for surgical procedures involve an interplay between the sophisticated automated platform, on the one hand, and the surgeon, along with his/her team, on the other (Alemzadeh et al., 2016). The outcome of such shared task performance essentially depends on how they can be attuned to one another (Fosch-Villaronga et al., 2021).

**1.1.2 Autonomy levels**

Generally, robotic surgical systems operate within three different function areas of medical practice, namely: 1) acquisition and analysis of information, 2) division of surgical trajectories or plan of actions, and 3) execution of the surgery (Manzey et al., 2009). Current surgical robots used to assist a surgeon performing (specific functions of) surgical procedures have different degrees of autonomy, ranging from no autonomy to full autonomy, and passing by being under the control of or in cooperation with a trained practitioner (Fosch Villaronga et al., 2021). Unlike the automation levels for automobiles by the standard SAE J3016 established by the Society of Automotive Engineers (SAE), currently, there are no universal standards that define the levels of autonomy in surgical robots. Nonetheless, Yang et al. (2017) have proposed a five-layered model for medical robotics autonomy levels, which has been further extended and refined in the literature (Rosenberg, 1993; Varma & Eldridge, 2006; Yang et al., 2017; Ficuciello et al., 2019; Fosch Villaronga et al., 2021) (*see* figure 2).

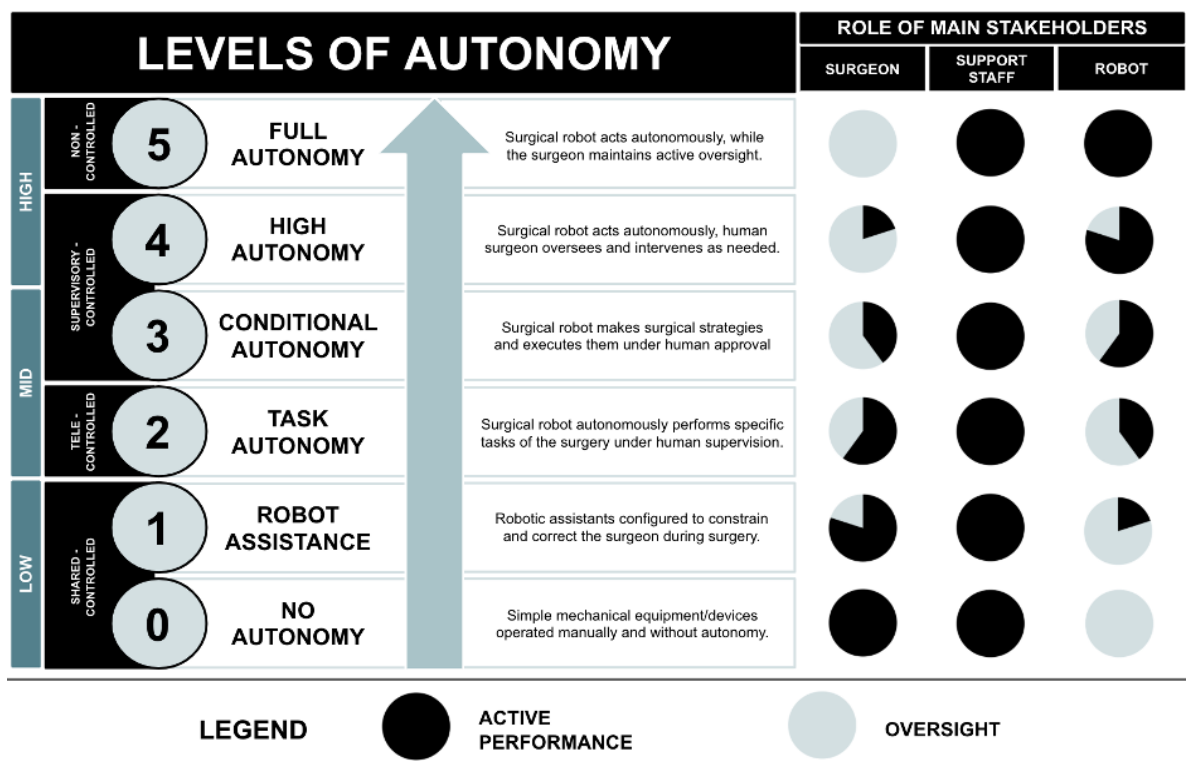

**Figure 2**. Autonomy levels and the role of humans in robot surgeries (Fosch-Villaronga, et al., 2021)

Recent developments in RAS have suggested a strong affinity towards increased autonomy levels amongst stakeholders, with level 4 autonomy surgical robots currently being in the development stage. Within this context, the term *autonomy* refers to 'the quality or state of being self-governing' (Merriam-Webster), and, as such, the term *robot autonomy* refers to 'a robot's capability to execute specific tasks based on current state and sensing without human intervention' (ISO 8373:2012). In practice, this means that surgical robots can perform certain functions of the surgical procedure independently. At the same time, humans are still needed for other parts of the process, including supervision, thus not leading to the complete elimination of humans (Fosch Villaronga et al., 2021).

Based on the robots' capability and the surgeon's role in performing the desired task, surgical robots can, generally, be classified into three categories: 1) shared-controlled, 2) tele-controlled, and 3) supervisory-controlled (Boyraz et al., 2019). The difference in these approaches held by surgical robots assisting in RAS primarily revolves around the robot's autonomy level, the degree of assistance provided by robotic systems during the execution of surgical procedures, and the human surgeon's control exercised.

The *shared-controlled* approach refers to a surgical environment in which one or more robotic devices work in tandem with the surgeon, meaning that the surgeon carries out the procedure with the use of a robot that offers steady-hand manipulations of the instrument, enabling the surgeon and the robotic system to jointly perform the surgical procedure in question (Mohammad, 2013). The shared-controlled approach follows the method by which the workspace is divided into several segments. The robotic device behaves differently based on different localization - safe, close, boundary, or forbidden classifications - and driven by haptic feedback (Boyraz et al., 2019) that will grow stronger as the surgeon's cutting tool comes closer to fragile human tissue, pointing out to the surgeon that extra caution ought to be taken.

The *tele-controlled* approach (also: master-slave, remote-controlled, or telesurgical approach) allows a human surgeon to operate the robotic surgical device from a (close) distance with no pre-programmed or autonomous elements. In 1996, the first FDA-approved robotic surgical system, ZEUS, was introduced, a complete robotic surgical system with seven degrees of freedom, tremor elimination and motion scaling (Ranev & Teixeira, 2020; Zemmar, Lozano, & Nelson, 2020). ZEUS was used for the first long-distance telesurgical procedure. Another breakthrough within this context was the da Vinci robotic system by Intuitive Surgery, which is used across different surgical specialties for a variety of surgical procedures and is capable of performing technically challenging procedures (Marescaux, J. et al., 2001; Troccaz, Dagnino & Yang, 2019; Zemmar, Lozano, & Nelson, 2020). Generally, a tele-controlled surgical robotic platform consists of one or more robotic arms (slave element), to which the surgical instruments are attached using a console (master controller), and which are generally configured with an optical system and computer-aided motion stabilization with a plurality of sensors for providing haptic feedback to the surgeon. The telesurgical approach requires the surgeon to manipulate the robotic arms - often characteristic to surgical robots, and telesurgical robotic systems in particular - during the procedure rather than allowing the robotic arms to work according to a predetermined program (Mohammad, 2013). Using real-time image feedback, the surgeon performing the surgical procedure can operate remotely using sensor data derived from the robot (Mohammad, 2013). As such, the robotic arms form an extension of the surgeon's actual hands. This is in line with the original intention of robotic surgery, which was to permit the performance of a surgical procedure from a remote distance without touching the patient (Satava, 2002). Since the outbreak of the COVID-19 pandemic, and the resulting unprecedented demands for hospitals, this vision has gained increasing importance. To effectively reduce pathogen spread, robots have been integrated in several sections of the surgical sequence, which each surgical patient traverses during a hospital stay. This workflow can be divided into pre-operative care, anaesthesia, the surgical procedure and postoperative care, and the overall goal of robotic integration is to minimize contact between the patient and healthcare provider at each step (Zemmar, Lozano, & Nelson, 2020). Digitization and machine intelligence have thus been called to action in the healthcare environment to combat the virus, and it is widely believed that their legacy may well outlast the pandemic and revolutionize surgical performance and management altogether (Zemmar, Lozano, & Nelson, 2020).

The *supervisory-controlled* approach is the most automated of the three methods, and the robotic platforms following this approach generally comprise multiple robotic arms equipped with different surgical tools. Moreover, they are often powered by AI (e.g., Artificial Neural Networks (ANN) and fuzzy logic (Amato et al., 2013). The supervisory-controlled approach entails robotic systems configured to perform certain functions of the surgical procedure for a large part - although not yet fully - autonomously, with the surgeon being in a supervisory role throughout these moments of the process. In other words, the surgeon remains indispensable in devising a surgical strategy and overseeing the execution of the robot (Mohammad, 2013). This thus means that the robotic devices are not yet capable of performing the concerning surgery without human guidance, as surgeons are often required to complete extensive preparations prior to the execution of the surgery and supervise during the execution thereof. As illustrated in figure 2, the human surgeon's role changes from active performance to supervision. Likewise, the surgical robot's role transitions from supervision to active performance as the surgical robot's autonomy increases. However, while the surgeon thus remains indispensable in preparing and planning the surgical procedure and overseeing the execution thereof, he no longer partakes in the surgery's execution directly (Mohammad, 2013; Fosch Villaronga et al., 2021).

## 1.2. Considerations for healthcare robot policymaking

### 1.2.1. The gradually transitioning position of surgical robots on the autonomy ladder

Research has pointed out that while the majority of RAS deploys surgical robots that follow the tele-controlled and shared-controlled approach, supervisory-controlled and fully autonomous surgical robotic devices have not yet found their way into RAS. Only performed on pigs so far, robot surgeons cannot perform an entire surgery completely independently from the beginning until the end on humans yet (Shademan et al., 2016; Greenemeier, 2020). Surgery is not only about enhanced dexterity but also about context understanding. Until now, human surgeons have shown to be considerably better than robots at weighing their experience to understand a particular context and make complex surgical judgments. In this sense, like cruise control and park assistance have made their way into cars progressively before realizing fully autonomous driving, fully autonomous surgical devices will gradually enter clinical practice (Svoboda, 2019). Although commercially available robots with level 5 autonomy, therefore, seem distant, some researchers, nevertheless, conceptualize them. Above all, the efforts to move from the presently available level 3 robots towards level 4 robots indeed suggests that, in principle, the deployment of surgical robots with fully autonomous capabilities equivalent to level 5 is the ulterior motive of researchers and engineers working in this field (Yip & Das, 2017). In this sense, as is the case with the current and continuously progressing autonomy transitions experienced within the field of autonomous driving and autonomous vehicles, fully autonomous surgical devices are likely to enter clinical practice (Svoboda, 2019), albeit in the distant future.

### 1.2.2. An increase of autonomy does not imply a decrease of human oversight

While Yang et al. (2017) state that the more autonomous medical robots become, the less human oversight exists, the phrasing *oversight* may give the impression that RAS is gearing towards humanless surgeries, while this is not the case (*see* figure 2) (Fosch-Villaronga et al., 2021). Even though surgery robots operate increasingly autonomously, this does not mean that humans are completely out of the loop during surgical procedures. Humans will still perform many tasks and play an essential role in determining the robot's course of operation. For instance, the medical support staff's role will remain integral and crucial to the surgical environments (e.g., for selecting the process parameters or for the positioning of the patient). Even the most autonomously performed surgeries, thus, still require humans for several actions during the operation. Among which (but not limited to) the provision of patient pre-operative care, the task of maintaining patients in a stable condition before the surgery, and the task of selecting the surgical robot's trajectory and task to execute throughout the relevant surgery, configuring the surgical robot (Crouch, 2020), and positioning the patient in the correct way relative to the robot.

Thus, the medical staff will maintain an integral and crucial role within the surgical environment determining the robot's course of operation. However, despite having a well-defined function in place, the surgery robot may unintendedly cause harm when it behaves in a way that differs from the designer's intent (Pistono and Yampolskiy, 2016). While this nuance is of importance to avoid excessively ascribing or extending responsibility to the surgical robot, it is also important to note that this does not mean that we have technology in place without human responsibility (Johnson, 2015; Bryson, Diamantis, & Grant, 2017; Fosch-Villaronga, 2019). Understanding the exact surgeon's role in RAS is essential to understand better who is responsible if something goes wrong.

### 1.2.3. User group and surgery type specific developments in the context of RAS

Based on the data compiled throughout this research, RAS does not seem overly male or female-centered. The benefits RAS provides within the field of MIS appears to be equally distributed among males and females. While RAS in many respects is more beneficial compared to open surgical procedures and has become the first-choice surgical modality for several surgical procedures - with the most common being prostatectomy and hysterectomy (McDermott et al., 2020) - research has nevertheless shown that patient's and medical staff's RAS perceptions often do not accurately reflect reality. Research identifies that females generally express more concerns regarding the safety and perception of new technology in surgery than males, who tend to be more unfazed by the notion of robotic surgery. This is partly due to the different ways in which males and females have shown to understand and conceptualize the process of RAS - where males tend to humanize this process; female participants consider it to be dehumanizing (McDermott et al., 2020).

Furthermore, pediatric surgical practice has not widely embraced robotics yet. The first reason lies in standard robot sizes, which usually hampers RAS's adequate performance on younger pediatric patients, although some hardware progress has been identified (Mehta & Duvvuri, 2012). While RAS may contribute to a decreased length of hospital stays and pain medication use, they are also associated with longer operative times. Their purchase and maintenance are very costly, which adversely affects pediatric hospitals' tight financial margins (Geller & Matthews, 2013; Howe, Kozel, & Palmer, 2017).

Additionally, esthetic surgery has so far not been accommodated to robot technology. In 2012, the literature did not reveal articles describing robotic surgery's use for cosmetic surgery (Ibrahim et al., 2012). The lack of robots in cosmetic surgery may respond to the fact that this type of surgery relies mostly on the surgeon's aesthetic abilities rather than technical precision and mechanical execution (Ibrahim et al., 2016). Cosmetic surgery depends very much on tactile feedback: from the preoperative assessment, marking and assessing the tissues and sutures tension are mostly done by the surgeon's hands to sense contour deformities and irregularities. Tactile feedback is also crucial during the surgical procedure itself, although current RAS rely on visual cues.

Likewise, so far, robot technology has very slowly and cautiously been embraced within dental surgery, with only a few clinical studies at present. However, the cost-benefit and cost-effectiveness of robotic surgery in implant dentistry are significant, and early experience with robotic dental implant surgery accuracy for implant osteotomy preparation is beginning to gain traction (Wu et al., 2019). In the United States, the first robotic dental surgery system, Yomi (Neocis Inc, Miami, FL, USA), was cleared by the FDA for dental implant procedures in 2017. This first system is used for dental implant placement and provides software for planning, gives navigational guidance for instrumentation during implant surgery, delivers haptic feedback, and controls the position, depth, and angulation for implant osteotomy. Furthermore, at the end of 2017, the world's first autonomous dental implant placement system was introduced (by Zhao) (Haidar, 2017). In addition to its high degree of autonomy, this intelligent robot can automatically and continuously adjust during intraoperative procedures. It can execute surgical tasks directly on patients without any apparent control by a surgeon. While novel within and promising for the field of robotic surgery, only limited confirming research is available regarding the reliability and feasibility of this system in clinical practice. More research is necessary to evaluate inaccuracy and implant positioning to validate robot intelligence–generated procedures (Yiqun et al., 2019).

One of the determining factors that influence implant therapy outcome and associated rehabilitation is implant placement accuracy (Yiqun et al., 2019). Although surgical navigation systems and template guidance meet these demands of high accuracy and surgeons have tried to use this technique to reduce errors of implant positioning, the physical position of a surgeon is often constrained due to the limitation of a patient's mouth opening and the location of missing teeth (Yiqun et al., 2019). Due to this limitation, the surgeon's performance may be affected, and human error cannot be eliminated. This may be solved through robotic surgery, which has the advantage of sustained precision, increased stability, greater efficiency, and more flexibility for assisting dental implant preparation and implantation. Despite its limitations and early development difficulties, robotic use in dental surgery seems promising as systems improve and reduce costs.

### 1.2.4. The inadequacy of surgery robotics training

While robotic surgery has shown to have a steep learning curve (Liu & Curet, 2015), there is no agreement on a specific learning curve, leading to disparities in the number of training hours among hospitals. As a result, it is uncertain what satisfactory certification is required to be considered a robotic surgeon (Ferrarese et al., 2016). While robotic surgery has exponentially increased over the last decade, a lag has been acknowledged in the development of a comprehensive training and credentialing framework. There are currently no standardized training modules for the use of surgical robots (Guzzo, & Gonzalgo, 2009). Subsequently, it is uncertain what satisfactory certification is required to be considered a robotic surgeon. Moreover, the currently available external training programs are limited to merely familiarizing the practitioner with the surgical robot's correct usage. Surgeons and practitioners must acquire the necessary skills to use surgical robots through external training, which ultimately depends on the quality of the robot manufacturer's training and may vary from one region to another (Fosch-Villaronga et al., 2021).

The incorrect use of medical equipment is a major reason for medical errors (Ward & Clarkson, 2004; Courdier et al., 2009; Hempel et al., 2015). Not having a clear framework that establishes a common minimum baseline for robotic surgeons creates incongruity in procedural safety. In this respect, a sound basis for such a framework demands a clear and elaborate overview of the most significant categories currently established within the field of surgical robots, their intended purpose, use, and main characteristics. In addition, to master robotic surgical techniques, surgeons need to acquire a set of core skills independent of the skills necessary to perform surgical procedures. Given that even in the most advanced RAS, surgeons and their teams still perform multiple functions, an optimal training framework on the use of surgery robots must be established (Fosch-Villaronga et al., 2021). The need for formal assessment of competency to ensure safe and sustained growth has led to various proposals for competency-based training programs in RAS (Sridhar et al., 2017). Generally, such training programs should entail technical training tailored to the surgical robots' type and autonomy level. They must include both 1) dry lab training on machines such as, but not limited to, Virtual Reality (VR) simulators that can simulate real-time situations and challenges and can be used for practice; and 2) wet lab training through real human/animal robotic surgeries which teaches the 'reaction of tissues' when the robotic instrument operates, such as dissection, excision, or suturing (Fosch-Villaronga et al., 2021). In RAS, most systems classify as master-slave systems in which the console is the interface controlling mechanical movement (Narula & Melvin, 2007). For RAS procedures to unfold smoothly and with little complications, similar to any advanced technological training, knowledge and working of the console is of paramount importance. Proficiency in basic console skills (such as camera, pedal, finger control) can be achieved in a relatively straightforward manner in a dry lab or VR simulated environment, in which individual and team reaction to system errors can be simulated, repeated and assessed (Sridhar et al., 2017). The development of non-technical skills to function in this complex multidisciplinary environment is an integral part of training and should best occur in parallel with developing the aforementioned technical skills (Sridhar et al., 2017).

Non-technical skills that are considered to significantly impact surgical success and that can be developed easily in a simulated environment include teamwork, leadership, situational awareness, and decision-making (Yule et al., 2006). The importance of having a validated training curriculum in place not only springs from the responsibility towards patient safety but also the ensuing issues with credentialing and associated liability (Sridhar et al., 2017). Although proficiency-based training curricula that comprehensively address the skills necessary to perform robotic operations (e.g., the da Vinci® Training Modules and materials, and the Fundamentals of Robotic Surgery (FRS) program) have shown construct and content validity as well as feasibility and measurable improvement in skills, they are limited in number, lack uniformity in credentialing and oftentimes do not account for the development of non-technical/team skills.

## 2. Assistive Robots

Assistive robots are service robots capable of assisting users (Tanaka et al., 2015). However, the concept of assistance typically anchors the physical and the practical: carrying a heavy load with an exoskeleton, performing exact surgical movements, or executing a menial task. Thus, assistive robotics typically refers to robots that give physical support to people with physical disabilities (i.e., *physically assistive robots*) (Miller, 1998). However, an all-embracing and updated definition should also cover *socially assistive robot*s, i.e., those robots that assist users through non-physical interaction (Feil-Seifer, & Mataric, 2005) using social cues with older adults in nursing homes like the flowerpot robot Tessa. In this sense, Nejat, Sun, & Nies (2009) define *assistive robots* as robots 'aiding, assessing, and motivating those in need, including patients, the elderly, and individuals with disabilities.'

We define *assistive robots* as 'service robots assisting a user through physical or social interaction'. Unlike the notion of a care robot, which focuses on robots 'used in the care of persons in general' (van Wynsberghe, 2013), our concept of primary stakeholder assistive robots focuses on the receiver of medical attention without necessarily being limited by the scope of care as Vallor (2011) suggested. In that light, we differentiate between two different types of assistive robots: socially-assistive and physically-assistive. While the latter uses the robot embodiment as a tool or potential extension of human's physical capabilities, the former is more aligned with the concept of traditional caregiving. There is a gradient of physical interaction within these two categories, revealing the importance of robot embodiment. Physically-assistive robots assist either through accompanying and supporting the user performing a task, thus in constant contact with the user, or performing a task for the user and requiring little to no contact (Fosch-Villaronga, 2019). For socially-assistive robots, the gradient is more nuanced. There is a spectrum that spans from remote contact robots where the main goal is not the physical interaction, but assistance is provided through social cues, like NAO; to a mix of social/physical interaction, where robots use social cues and invite the user to have physical contact with them, such as Paro, where contact has been found to benefit the patient.

### 2.1 Socially-assistive robots

While social robots represent a shift towards highly interactive robots and entail deeper Human-Robot Interaction (HRI) (Breazeal, Dautenhahn, & Kanda, 2016), Socially Assistive Robots (SAR) intensify that process by providing direct support to users. Due to the broad definition of assistance (Merriam-Webster) and the almost infinite scope of social interaction, unlike surgical robots, the development of SARs cannot be aligned with or limited to a single purpose (Hegel et al., 2009; Li, Cabibihan, & Tan, 2011; Aymerich-Franch & Ferrer, 2020). The industry and promising research are not oriented towards the optimization of a single issue but rather navigate the numerous entanglements between the needs of the patient, the translation of those needs into concrete assistance, and how social interaction can modulate the assistance provided by robots.

When enmeshed within the development of new technology, the simultaneous fluidity and opaqueness of these entanglements lead to a confusing but fruitful development of devices that flirt the boundary between the categories of medical device, toy, and product (Fosch-Villaronga, 2019). The flexibility of this distinction regarding SARs requires a precise definition, categorization, and analysis to determine the future role and scope of action of the law.

SARs represent the intersection between assistive robotics and socially interactive robotics: SARs assist through social interaction (Seifer & Mataric, 2005). The field is primarily oriented towards developing robots capable of close and effective interactions to provide optimal assistance (Scassellati, Admoni, & Matarić, 2012). Unlike chatbots or other AI-driven assistive technologies, SARs specifically use their embodiment (arms, sensors, or touchscreens) to generate, modulate, and provide assistance through interaction. Typical embodiments include anthropomorphic, zoomorphic, caricatured, and functional (Fong, Nourbakhsh & Dautenhahn, 2003) and this embodiment plays a crucial role in many applications: children feel stronger friendship bonds with a physical robot compared to a virtual avatar, a physically present robot tutors produces better learning results, and individuals with cognitive impairments find the interaction more 'efficient, natural, and preferred' with a physical robot than with a simulated one (Tapus, Tapus & Mataric, 2009; Leyzberg et al., 2012; Sinoo et al., 2018). Robot embodiment enhances presence, helps with the allocation of social-interactional intelligence, typically via gaze and facial expressions, and makes robot task capabilities intelligible from the user perspective (Tanaka, Nakanishi, & Ishiguro, 2015), which may enhance the transparency of robot intentions and actions and promote trust (Felzmann et al., 2019). Although we acknowledge the rise of virtual conversational agents (Adiwardana & Luong, 2020) like chatbots and 'virtual robots,' these fall outside this study's scope.

### 2.1.1. SAR Categories

Compared to surgical robots, SAR categories are blurrier due to the range of services and functions (including the medical field) falling under the umbrella of assistance. The various nomenclatures used to call these robots and the fact that this subfield of robotics is in its infancy does not help. Libin and Libin (2005) differentiate between social, educational, recreational, rehabilitation, and therapy robots. Other authors group the first three categories under 'care robots' (Vallor, 2011; van Wynsberghe, 2013). However, although there is a clear subset of SARs for (cognitive and social) therapy, SARs often have blended applications where their primary use depends on both the context and the user. For example, robots such as AIBO are toy robots used in studies on dementia or loneliness (Tamura et al., 2004), and SoftBank Robotics' NAO is a social robot for which the only one of its many applications relates to assistance in a healthcare context.

We distinguish between care and therapy robots, depending on the type of assistance provided (*see* table 5). Care robots primarily provide social interaction and support in any environment, including a healthcare setting. For instance, keeping an elderly person company in their home. Therapy robots assist users with a specific form of therapy, which are condition and environment-specific, and thus used within controlled environments, which implies that a form of monitoring of the interaction between the user/patient and the robot is taking place. Some care robots are also used in monitored environments. This distinction, though highlighted in this paper, is not as clearly seen from the robot providers' perspective.

A general division of SARs into therapy robots and care robots provides a more explicit framework that accentuates the need for manufacturers to state such robots' *intended use* to avoid misclassification and lack of the necessary safeguards to ensure safe use. Not having a clear category (toy, medical device, product) challenges the application of a subsequent normative framework and questions the safeguards applied to ensure the safety of the robot itself and the HRI (Fosch-Villaronga, 2019). Distinguishing between care and therapy may have significant legal ramifications as therapy robots' controlled nature lends itself more readily to their classification as medical devices. Still, this effort should be accompanied by recognizing specific therapies, such as robot pet therapy, as accepted alternative forms of therapy, which currently do not enjoy a recognized intervention category as animal therapies (Fosch-Villaronga & Albo-Canals, 2019).

| Classification of Socially Assistive Robots | |
|---|---|
| **Category** | **Subcategory** |
| Therapy | - Dementia<br>- Autism<br>- Neuro-developmental disorders |
| Care | - Companion[2]<br>- Basic assistance<br>- Robot pet therapy [3]<br>- Aging-in-place in EU also called *Active Assisted Living*[4]<br>- Sex care robots |

**Table 5**. Classification of Socially Assistive Robots

**Therapy Robots**

Therapy or therapeutic robots are commonly defined as robots used for *robotherapy*, a framework of HRI through which a series of coping skills are developed and mediated through robots (Libin & Libin, 2003). The notion of robotherapy, like that of assistive robots, is, in practice, more oriented towards the physical (and more specifically, rehabilitation) than cognitive or psychological (Krebs & Hogan, 2006). Therapy robots exist both in socially-assistive and physically-assistive forms (Lorenz, Weiss & Hirche, 2015). Socially-assistive robotherapy is any form of psychological or cognitive therapy mediated through robots and, more specifically, through social robotic interaction (Libin & Libin, 2005). These robots are usually used in already-existing therapies and serve a well-defined purpose (Rabbitt, Kazdin, & Scassellati, 2015).

**Care robots**

Like the notion of assistance, that of care is vague and multiple. Assistance may be a form of care, and care may be a form of assistance. Care robots represent a prolific research domain in social robotics, yet care robots, assistive robots, and socially-assistive robots are treated in the literature as distinct, independent categories without any relations of interdependence or entailment. The ISO 13482:2014 even goes further and establishes the category 'personal care robots.' While ISO 13482:2014 does not define *personal care* (Fosch-Villaronga, 2016), it defines such robots as 'service robots that perform actions contributing directly towards improvement in humans' quality of life, excluding medical applications.' Within the context of socially-assistive robots, care robots are robots whose mode of assistance improves humans' quality of life via socially-robotic interaction (in the ISO 13482:2014, *mobile servant robots*). Given the growing literature exploring new uses for sex robots in the care sphere (Jekker, 2020; Fosch-Villaronga & Poulsen, 2020), SARs also include sex robots.

---

[2] Companion robots are also promoted as mental health robots as they lessen loneliness through the provision of robotic companionship.

[3] Robot pet therapy is like robotherapy in the sense that it is defined simply as therapy with the medium of an animal. Most animal therapies are not diagnosis-specific but focus on alleviating side effects like loneliness.

[4] Category coined by Lorenz, Weiss, and Hirche (2015). Comprise robots that facilitate care of the elderly (assist with tasks, remind when to take medicine etc.) and fall within the scope of healthcare services.
*Also see* http://www.aal-europe.eu/.

Sex robots are *service robots that perform actions contributing directly towards improvement in the satisfaction of the sexual needs of a user* (Fosch-Villaronga & Poulsen, 2020; 2021). The potential use of sex robots for disability care purposes includes the physical and mentally ill. Their raison d'être is simple: although every human should enjoy physical touch, intimacy, and sexual pleasure, persons with disabilities are often not in the position to fully experience the joys of life in the same manner as abled people (Fosch-Villaronga & Poulsen, 2021). Although there have been international efforts towards realizing their sexual rights from institutions like the United Nations (1993; 2007) after nearly 30 years of discussion, this topic remains an unfinished agenda for the disabled (Temmerman, Khosla, & Say, 2014) as if contemporary societies succeeded in accepting (at the time) highly controversial social phenomena like same-sex marriage and transgender people, but they failed to recognize people with disabilities as sexual beings (Maxwell, Belser, & David, 2007; Roussel, 2013).

Typically, sex robots have different embodiments, including fully or partially-bodied humanoids; body parts such as arms, heads, or genitals used for sex-related tasks; or non-biomimetic robotic devices used for sexual pleasure. These robots usually display realistic sex-related body movements, have sensors to react in real-time to user interaction, and include human cues such as voice, gaze, and lipsync to support human-like HRI. Depending on their embodiment, sex care robots could be also classified as physically assistive robots. Beyond satisfying sexual pleasures, these robots may help address first-time sex-related anxiety, treat sexual dysfunctions, or promote safer sex in educational settings. Sex robots could create a safe, non-judgmental environment for people who feel insecure about their sexual orientation (Levy, 2009; Royakkers & van Est, 2015). Other more controversial applications are treatments for pedophiles and potential sex offenders (Danaher, 2017).

**2.1.2. Considerations for healthcare robot policymaking**

*2.1.2.1. Definitional opposition and blurriness in Socially Assistive Robotics*

SARs state of the art illustrates the definitional opposition between the robot categories of care and therapy. One-on-one correspondences between the intended purpose and context of use are only visible for the robots falling under the therapy category. There is a clearly outlined domain of interest for those robots, making for coherence between the company's goal and the robot's day-to-day application.

The very existence of care as a domain of application of healthcare technologies instead allows for the blurriness in the robots' application, notably revealing an unclear boundary between robots for serviceable contexts and healthcare robots. This is notably seen with NAO and Pepper, where healthcare is just another domain of application or vertical and not a field in its own right. Essentially, any social robot can be used in medical, therapeutic, or educational settings, effectively trespassing the categories of medical and non-medical devices. This trespassing becomes evident in robots such as PARO, where the robot, though designed to be used in a medical, therapeutic context, has FDA approval in the US and the CE mark in the EU (Shibata, 2012). In other contexts, toy robots may be used for medical or therapeutic uses without being considered as medical devices. These blurred contexts of use make it increasingly difficult to define what counts as a medical device according to the European medical device regulation. More importantly, it stresses the need to oversee these in-between cases, such as My Special Aflac Duck,[5] used in healthcare settings, despite not being considered medical devices. These cases may present unforeseen risks to a user that could have been prevented by following the medical device regulation. For instance, Pleo is a toy robot used in an experiment studying HRI with autistic children (Peca et al., 2014) and as an emotional support tool for the elderly in nursing homes (Baisch et al., 2018). This pattern of commercial non-specialized robots being used as companions to fulfill emotional and psychological needs reveals the general tendency for mental health and therapy to be subsumed in the category of companionship (Hutson et al., 2011).

*2.1.2.2. Underdevelopment of autism research in robots with youth and adults*

We notice that the target end-users for most SARs' applications are the elderly and children. However, when it comes to children, the focus is primarily on the autism spectrum disorder, a disproportion which, in turn, reflects the tendency for autism research in robots to be conducted exclusively with children (Scassellati, Admoni & Matarić, 2012). Our work shows the significant lack of studies observing HRI with autistic youth and adults. The large underdevelopment of this field of research may be a byproduct of the tendency for SARs to be presented or marketed as toys. This overfocus on children as the main subjects of studies using SARs in various treatments for autism then has repercussions on the way therapy is approached for adults, highlighting the infantilizing tendencies of care (Salari & Rich, 2001; Marson & Powell, 2014). Ideally, SARs' embodiments need to adjust to a wide range of individuals who will come in contact with them and cannot be limited to the narrow pool of subjects determined by research.

*2.1.2.3. Discrepancies in usefulness and repurposing of Socially Assistive Robotics*

The use of technology in mental health is somewhat scattered in service of the exploration of robotics and AI applications. Although research on robots for depression (Chen, Jones, & Moyle, 2018), to provide companionship, and to provide specific therapies, such as Cognitive Behavioural Therapy, is growing (Costescu, Vanderborght, & David, 2017), research does not explicitly show how the majority of work in the field of SARs is conducted in experimental contexts. Unlike surgical robots, there is a large discrepancy between commercially available robots and the different robot applications and uses researched in labs but that never make it outside of experimental contexts. The majority of novel robot applications, like Bandit or Charlie, are developed within the context of projects or labs and designed for specific research purposes. In reality, very few robots, whether commercial or experimental, are actively used by users in their daily lives.

The limited scope of application highlights another problem within SAR research. Because of the sheer lack of robots, there is substantial repurposing. Studies use robots commercialized as toy robots in therapeutic settings to determine if they can be used in novel ways. These experimental settings favor testing the use of already established robots such as Paro or NAO in novel situations because of their widespread availability. For example, Scoglio et al. (2019) find that NAO is one of the most used social robots in mental health and well-being studies. It is also used in studies for elderly care of persons with Chronic Obstructive Pulmonary Disease and health education for children with diabetes (Blanson Henkemans et al., 2013).

---

[5] *See* https://www.aflacchildhoodcancer.org/.

The latest findings point out to two opposite directions concerning SAR usefulness (Fosch-Villaronga & Albo-Canals, 2019): while some researchers argue that by allowing the robot to show attention, care, and concern for the user as well as to being able to engage in genuine, meaningful interactions, socially assistive robots can be useful as therapeutic tools (Turkle, 2007; Shukla et al., 2015); other studies suggest that the emotional sharing from the robot to the user does not necessarily imply feeling closer to the robot (Petisca, Dias & Paiva, 2015). Moreover, the additional moral question of whether robots such as Paro are suitable as companions replacing human interaction is growing in the literature (Calo et al., 2011; Sharkey & Sharkey, 2012) and policymaking (European Parliament, 2017).

*2.1.2.4. Physical embodiment and human biases*

As opposed to surgery robots, which typically have a functional embodiment, SARs do not have an embodiment centered around functionality but instead prioritize aspects such as friendliness, safety, and ease of maintenance and cleaning. What is notable regarding SARs' embodiment is the tendency towards child-friendly, if not child-like, and toylike, appearances. Most available SARs navigate between anthropomorphic or zoomorphic forms. Milo, Kaspar, and also Nao and Pepper, among others, resemble a small child, while animal-inspired SARs, such as Pleo or Paro, are designed to evoke stuffed animals or toys. These embodiments largely replicate the relationships we establish with social robots, that of caregiver-infant or pet-owner (Pillinger, 2019).

Excluding their philosophical implications, anthropomorphic robots' byproduct consequence is the potential exacerbation of human biases, notably that of race and gender. Not only is the color white prevalent in most robot designs, but people have been shown to project racial biases onto black robots (Addison, Bartneck & Yogeeswaran, 2019; Bartneck et al., 2018). Additionally, the default for anthropomorphic robots tends to be white males. For example, Charles, Flobi, or Bandit, although Milo, an educational robot for autistic children, is offered in different skin tones. The relationship between biases and embodiment is thus complicated.

On the one hand, recognizably gendered robots lead to the projection of gendered stereotypes onto those robots (Eyssel and Hegel, 2012) while, on the other hand, having robots without recognizable human traits, such as gender, could potentially further the intrinsic dehumanization of a non-human agent taking over the human role of care (Calo et al., 2011). Furthermore, there is a projection of existing gender norms even onto non-human robot forms (Nomura, 2017), which shows that biases such as gender are so deeply embedded in our way of navigating the world that they cannot be ignored. At best, designers can consider following initiatives such as 'Gendered Innovations' from Stanford University[6] or the European Commission (2020b; Nature, 2020), or move beyond offering customizable robots into genderless robots and even beyond the anthropomorphic into the realm of the 'technomorphic' (Lütkebohle et al., 2010). Moreover, while robots with human or animal-like features are designed with the goal of being trustworthy, sociable and encouraging users to bond with them (Mende et al., 2019), this may nevertheless appear to be a double-edged sword: the more anthropomorphic the robot, the more likely the user will start to expect real empathetic care, which the robot cannot provide (Sheridan, 2016). A solution to this is to design humanoid robots with a visible demarcation between human and robotic features; for instance, designing the robot to have wheels instead of legs (Cresswell, Cunningham-Burley & Sheikh, 2018).

---

[6] *See* https://genderedinnovations.stanford.edu/.

Apart from a tendency towards children-friendly forms and toylike appearances, with regards to the embodiment, we can see that, very much like surgery robots, there is not much internal variation. SARs' functionality depends more on the software included and the services contracted (e.g., speech recognition software). Although the embodiment plays an important role compared to virtual agents, the AI and the software, together with the caregiver's therapy, make them different in every setting.

*2.1.2.5. Sex robots as a tool to help realize the sexual rights of persons with disabilities*

The full realization of the sexual rights of persons with disabilities requires more research and policies that understand the intersection of people, disability, and sexual rights (Addlakha, Price & Heidari, 2017). In Jecker's (2020) words, 'just as society has the power to insult people's dignity by shaming and stigmatizing their sexual desires and behavior, it has the power to support dignity and serve as a bulwark against shame.' These policies could represent a step forward in treating people with disabilities in a dignity-respecting and non-discriminatory fashion concerning their sexual rights (Fosch-Villaronga & Poulsen, 2021).

We anticipate, however, that technology increasingly and paradoxically has a habit of being both the source of and the solution to societal problems, and robots are no exception (Bauman, 2013; Johntson, 2018). Sex robots may offer a good solution to support human dignity for disabled populations, but these robots may also bring other unexpected consequences (Fosch-Villaronga & Poulsen, 2021). These reflections anticipate that the design and implementation of sex robots in care is not straightforward and requires careful thought.

*2.1.2.6. The 'physical' component of safety may not be the only dimension at stake in SAR*

Traditionally, the definition of safety has been interpreted to exclusively apply to risks that have a physical impact on persons' safety, such as, among others, mechanical or chemical risks (Fosch-Villaronga & Mahler, 2021). However, the current understanding is that the integration of AI in cyber-physical systems such as robots, thus increasing interconnectivity with several devices and cloud services, and influencing the growing human-robot interaction challenges how safety is currently conceptualised rather narrowly (Martinetti et al., 2021). An overfocus on physical safety conveys the impression that other aspects such as privacy, data protection, autonomy, psychological harms, or dignity do not play a role in ensuring a safe human–robot interaction (Fosch-Villaronga & Mahler, 2021). Thus, to address safety comprehensively, robotics demands a broader understanding of safety, extending beyond physical interaction, but covering aspects such as cybersecurity, and mental health (Martinetti et al., 2021).

## 2.2. Physically assistive robots

Physically assistive robots (PAR) are service robots that support users through physical interaction. PARs may assist the robot operator, the beneficiary, or both, through physical interaction to perform tasks (Canal et al., 2017). This interaction is not limited to specific contexts and thus includes the medical and rehabilitation. This physical assistance can be partial, where the robot acts as a supportive presence; or total, where the robot performs an action for the user.

In recent years, demand for physical therapy services worldwide has increased, partly due to aging populations. As a result, assistive technologies and rehabilitation robotics have become popular, especially as they promise to ease the stress on medical and physiotherapy staff and control expenses while simultaneously improving the lives of physically, cognitively, or neurologically impaired (PCN-impaired) individuals. Worldwide, many people suffer from chronic physical, neurological, and cognitive disabilities (WHO, 2011). PARs are increasingly used to improve persons' independence and quality of life with physical (Brose et al., 2010), cognitive, or neurological disabilities. These advances' clinical impact is observed as users can enter the workforce, lessen the burden to their caregivers, and live at home instead of in long-term care facilities, as medical complications are prevented, and self-image and life satisfaction are improved (Brose et al., 2010). Robotic systems used for this purpose vary from feeding robots to smart-powered wheelchairs and independent mobile robots to human-robot collaborative units.

The provision of physical assistance is one of the most direct ways that robots can help PCN-impaired persons. They allow for increased independence and function in physical tasks. ISO 13482:2014 defines a PAR as a 'personal care robot that physically assists a user to perform required tasks by providing supplementation or augmentation of personal capabilities.' Though the ISO standard provides a good basis for a definition and accounts for the need to have specific requirements for different user categories, it fails to consider concrete users such as children, elderly persons, and pregnant women (Søraa & Fosch-Villaronga, 2020; Fosch-Villaronga, Čartolovni & Pierce, 2020). This incompleteness that persists in the reviewed and confirmed version of 2020 brings about uncertainties about exactly what falls within the framework's protected scope.

### 2.2.1 Categories

Due to the broad scope of their use, there are many different types of PARs. PARs interact with humans and can be directly worn by them. Generally, within the field of PARs, a distinction can be made between restraint and restraint-free PARs (ISO 13482:2014). *Restraint PARs* are fastened to the human body during use and directly assist PCN-impaired persons by being attached to them. These are usually for the lower or the upper-limbs. *Restraint-free PARs*, on the other hand, are not fastened to the human body during use and, therefore, indirectly assist physically impaired persons. Within the context of this distinction, ISO 13482:2014 explicitly differentiates supplementation and augmentation. While *supplementation* refers to the assistance that restores an average level of human capability to persons who may otherwise have difficulty doing so due to their disability, *augmentation* refers to the physical assistance in the performance of physical tasks that exceeds what can be generally expected without assistance (ISO 13482:2014). Both the restraint/restraint-free distinction and the differentiation between supplementation and augmentation revolve around the concept of varying degrees of assistance, how they modulate the depth of the HRI, and the user's capabilities that result from it. Understanding these degrees of assistance is crucial when trying to understand their implications for robot automation and autonomy. Due to their interaction with the user, there is a tendency for restraint PARs to provide supplemented assistance. At the same time, restraint-free PARs tend to provide augmented assistance.

| Classification of Physically Assistive Robots | |
|---|---|
| **Category** | **Subcategory** |
| User Support | - Exoskeletons and Exosuits<br>- Prosthetics<br>- Robotic arms [7]<br>- Walking aid (walkers, rollators)<br>- Sensory-assistive robots (Hersh, 2015) |
| Task Performance | - Feeding robots<br>- Robotic manipulators<br>- Smart wheelchairs<br>- Robotic nursing assistants |
| User Rehabilitation | - Orthoses[8]<br>- End-effector robots<br>- Exoskeletons[9] |
| Body Part Replacement | - Robotic prostheses |

**Table 6**. Classification of Physically Assistive Robots

Since PARs are the industry figurehead for assistive robotics, different types of assistive robots are easily distinguishable as they are developed and marketed to solve a specific issue and target a specific market. The main difficulty lies in establishing overarching categories that unify the different needs and problems solved by each sector's robot applications and consider existing distinctions such as the one established by ISO 13482:2014. As such, we chose to distinguish PARs based on the depth and complexity of their assistance. Within the context of our classification, we distinguish PARs that support the user(s) in the performance of specific tasks, PARs that perform the physical task for its user(s), and PARs used for rehabilitation (*see* table 6). In addition to our overarching categorization of support, performance, and rehabilitation, the industry differentiates PARs based on the focal area of their assistance. Therefore, PARs that assist physically and neurologically impaired persons with specific tasks qualify either as upper or lower limb assistive.

**2.2.1.1. User Support**

Physically-assistive robots that solely support PCN-impaired individuals in specific tasks' performance focus on applications that allow a person with a severe disability to perform ADL and vocational support tasks that would otherwise require a human attendant (Brose et al., 2010). Such robots are under the control of their operator and are typically used to handle books, medication, paper, computer media, food and drink, control communication devices, and activate electrical appliances. Such robots' primary users are typically individuals who suffer from a severe physical or neurological disability that seriously limits their upper or lower limb mobility but can communicate clearly and have an average cognitive ability. Such disabilities usually materialize due to high-level Spinal Cord Injury (SCI), cerebral palsy, muscular dystrophy, and, more generally, for anyone who cannot manipulate household objects (Siciliano & Khatib, 2016). The increasing importance and reach of supportive PARs are reflected in the full range of supportive PARs currently commercially available worldwide (e.g., feeding and drinking robots, bodyweight and body movement supportive robots, robotic arms, exoskeletons, and prosthetics).

---

[7] Not to be confused with exoskeletons as these are robotic arms that attach to wheelchairs or tabletops and support the user's movements. Some models include slings within which patients can place their wrists or elbows. Other models offering performance-type assistance have been collected in the category robotic manipulators. See for example iFLOAT Arm.

[8] Though robotic orthoses are often grouped under the label of exoskeletons, orthoses are medical devices to which Article 1.3 of the Medical Device Regulation (MDR) is applied as well as being defined in ISO 22523:2006.

[9] Exoskeletons are included twice in this table as their use for medical applications is more regulated and context-specific than that of personal care (support).

### 2.2.1.2. Task Performance

Performance PARs assist users beyond the realm of support and instead perform the required task for the user. Like support PARs, a significant proportion of these robots fall under the scope of care as they assist PCN-impaired individuals in the performance of ADL. For instance, feeding robots can promote independence and more intimacy during mealtimes (Herlant, 2018). These robots represent an end of the spectrum of robotic assistance where human control over the devices is indirect or, in the case of fully autonomous/automatic robots, non-existent. The degree of human-robot collaboration creates a range of autonomy in an action's performance, which is of increased importance when it comes to performance robots as it determines the degree of dependence the user has on the robot. This, in turn, has serious safety (Vasic & Billard, 2013) and, therefore, legal and medical implications (Hersh, 2015). Moreover, because of the growth of the elderly population, there is a search for optimizing systems of care by alleviating the routine tasks done by nurses. There is a rising trend in robot nurses, and the growing incorporation of robotic assistance into the healthcare ecosystem. These robots are designed and deployed to assist doctors in the hospital context in the same way that human nurses would. Such robots are currently already being used in Japanese hospitals, allowing the Japanese healthcare sector to adequately and efficiently treat their large elderly population - Japan has the highest percentage of elderly (75+) individuals among OECD countries. Even more so, Humanoid Nursing Robots (HNR) are believed to replace future nurses in Japanese healthcare facilities in the (near) future (Khan, Siddique & Lee, 2020). Moreover, the deployment of such robots relieves the nursing and other healthcare staff that generally undergo high stress and exhaustion due to patient load. This is a challenge especially highlighted since the outbreak of the COVID-19 pandemic. Paro by AIST, Pepper by Softbank Robotics, and Dinsow by CT Asia Robotics are a couple of examples of nursing robots used to assist elderly patients providing lifting as well as in therapeutic assistance. Within this context it is thus important to stress that task performance robots assist all primary stakeholders, not just receivers of care. ISO 13482:2014 names these robots 'Mobile Servant Robots' (MSR) if they incorporate a social interface. However, task performance PARs work differently from Healthcare Service Robots (these being closer to the MSR concept). For example, Robear is a nursing robot used to lift patients, Dinsow robot is used for elderly entertainment and face-to-face calls, and Moxi is a nursing robot that places medicine in bins.

### 2.2.1.3. User Rehabilitation

Rehabilitation robots are too often seen as constituting in and of themselves the field of robotherapy and perceived as distinct from assistive robotics as a whole. As seen for SAR and illustrated in the ISO standard, despite the growing body of blended applications, there still exists a definitional opposition, if not exclusion, of therapeutic physical robots within the industry. In essence, rehabilitation robots are PAR applied in therapeutic settings, but like

SAR, they are used in specific medical contexts under caregiver supervision and guidance. Rehabilitation robots are generally deployed to help in the rehabilitation of patients after an accident or stroke (Tsui & Yanco, 2007), to assist and treat the disabled, elderly, and inconvenient conditions of people (Khan, Siddique & Lee, 2020). They promote functional reorganization compensation and regeneration of the nervous system, effectively alleviating muscle atrophy (Zhao et al., 2020), thereby relaxing rehabilitation physicians and staff from their overwhelming physical labor and thus optimizing the available healthcare resources. Examples of such robots include the Kinova assistive robot by Kinova Robotics, and EksoNR by ekso bionics.

PARs also defy the label of personal care robots. As such, the assistance provided is limited in time (the timespan of the treatment) and working towards a specific goal (e.g., the development of neuroplasticity required to regain lost motor functions (Gassert & Dietz, 2018)). The distinction between upper and lower limb assistive robots is significant as therapies focus on one area of interest. Robots can further be classified within these different target areas based on the different degrees of cognitive and psychological engagement they require: grounded exoskeletons, grounded end-effectors, and wearable exoskeletons. (Gassert & Dietz, 2018). Interestingly, according to Fasoli et al. (2004), all PARs fall under the category of rehabilitation. In their paper, we dubbed support and performance robots as robots that compensate for lost skills instead of robots that remediate or retrain the area with lost skills.

### 2.2.1.4. Body Part Replacement

Medical implants are devices or tissues placed inside or on the body's surface usually intended to replace missing body parts, such as a limb, a heart, or a breast implant, which may be lost due to physical injury, disease, or congenital conditions (FDA, 2021). Medical implants can be prostheses or can have other functions such as delivering medication, monitoring body functions, or supporting organs and tissues. If they incorporate robotic parts, these are called robotic medical implants. The adjective robotic distinguishes passive adjustable devices from usually electrically powered mechatronic systems and stresses the presence of actuators, sensors, and microcontrollers and an intelligent control system implementing the desired behavior in these devices (Palmerini et al., 2014).

Some implants are made from skin, bone, other body tissues, and others from metal, plastic, ceramic, or other materials. These devices consist of three elements: 1) a biological (i.e., human or animal) part linked to 2) an artificial part (i.e., prosthesis, orthosis, or exoskeleton) using 3) a control interface (Micera et al., 2006). These systems also have different levels of hybridness, augmentation, invasiveness, and temporality (Palmerini et al., 2014; FDA, 2021):

- **Hybridness** refers to how close the artificial device and the human body are. These devices can be detached from the human body (i.e., teleoperated) or connected anatomically and functionally to the body, like a prosthesis.
- **Augmentation** or assistance level concerns the number, type, and degree of human capabilities empowered, restored, or supported.
- **Invasiveness** refers to how invasive the biological and artificial elements connect. The invasiveness level ranges from non-invasive, such as a joystick, or direct interfaces connected with the central or peripheral nervous systems, such as Brain-Computer

Interfaces (BCI) (Micera et al., 2006), which can be non-invasive such as EEG or invasive, such as implanted electrodes.
- **Temporality** refers to the fact that these systems can be placed permanently or removed once they are no longer needed.

Important to note within this context is that there is a fundamental difference between prostheses and orthoses. While prostheses entail replacing a missing body part or an organ, active orthoses improve the functionality of an existing body part (Palmerini et al., 2014).

### 2.2.2. Considerations for healthcare robot policymaking

*2.2.2.1. Gradience and blurriness within the field of Physically Assistive Robotics*

PARs can generally be divided into systems that support the user(s) in the performance of certain tasks, PARs that perform the physical task for its user(s), and PARs used for rehabilitation. Although each of these categories are strictly separate from one another and from other healthcare robot categories, in theory, they appear to be of a gradient nature in practice, often overlapping one another in their purpose and function. For example, there currently exists blurriness between robotic nurses as merely physical assistants and the growing desire for multipurpose and multi-functional robots, possibly leading to the existence of a mixed assistance category/trend heading in the direction of mixed assistance (Hersch, 2015). Robotic nurses encompass a complex set of purposes and uses, incorporating characteristics of both PARs (e.g. if they can lift patients), socially assistive robots (e.g. if they have a social interface) and healthcare service robots (e.g. if they only bring medicines). The establishment of fixed categories will likely prevent such complex robots from being properly assessed and addressed by policymakers.

Moreover, the presence of exoskeletons in categories of both support and rehabilitation illustrates the fact that contrary to the categorization of SARs, Healthcare Service Robots, or surgical robots, PARs do not have robot applications that can be neatly distinguished. Rather, the same devices or type of devices are differentiated solely based on use. This heightened importance of context of use is especially salient for lower-limb exoskeletons, where the medical and therapeutic environments in which these robots are used are often their sole differentiating factor. For example, *ReWalk* offers both a personal care robot and a stroke rehabilitation robot. Despite the overwhelming similarities in their functions, both applications are submitted to different regulations and thus legally are not treated the same. While the *ReWalk* exoskeleton is a personal exoskeleton, the *ReStore* Exo-suit is a medical device intended for use inside a clinic. This enforcement of rehabilitation robotics as being outside of the scope of the industry creates a false dichotomy between devices such as the *ReWalk* Exoskeleton and the ReStore Exosuit and further leads to a discrepancy in regulation and development with regards to devices of the same fundamental nature, where the safety of the user is defined and enforced in varying ways depending on the end goal of the producer. (Dellon & Matsuoka, 2007) The superfluousness of this distinction becomes even more apparent with the development of new orthotic devices which are actively breaking down the barriers between therapy and ADL in an attempt to have neurorehabilitation become more centered around functional activities. (Poli, Morone, Rosati & Masiero, 2013).

*2.2.2.2. Overfocus on adult populations, salient gap in access to children*

There is a broad range of PARs currently available to the public and used within an academic context. However, the introduction of robots for health-care and therapeutic purposes that interact directly with patients, older adults, and children has proved to be complicated. Within the field of PARs, generally, there is an overfocus on adult populations, while a particularly salient gap in access is found in children (Fosch-Villaronga, Čartolovni & Pierce, 2020). As mentioned at the outset of this section, the standard that regulates personal care robots (including physical-assistant robots and exoskeletons, ISO 13482:2014) fails to consider concrete users such as children, elderly persons, and pregnant women. While the standard recognizes that these populations deserve special consideration, no significant steps have been taken to realize such an imperative (Fosch-Villaronga, 2019).

For most PARs, especially exoskeletons, a determining factor within the context of their use is the height of the user involved. Generally, companies build exoskeletons for persons from 150 cm tall onward (Søraa & Fosch-Villaronga, 2020), while the average height of a 5-year-old in Europe is between 108 and 113 cm (Bonthuis et al., 2012). Company FAQ often clearly states the use population which the PAR addresses (e.g., Rewalk, Mealtime Partner Assistive Dining Device & Trexo & ReoAmbulator™ & Optimal-G™ Pro). This inevitably makes it particularly difficult to develop and create PARs suitable for both adults and children. Although advanced wearable exoskeletons are currently available, their benefits and challenges in addressing childhood movement disorders remain underexplored (Lerner, Damiano & Bulea, 2017). In this respect, the used height parameters prevent shorter populations from accessing the benefits of this technology.

Past research has shown that damage to the central nervous system during early development can lead to neuromotor disorders, such as cerebral palsy, that negatively affect users' quality of life, including children. Still, gait rehabilitation robotic devices primarily focus on the restoration of lost functions after stroke or spinal cord injury in adult populations rather than developing certain functions (Fosch-Villaronga, Čartolovni & Pierce, 2020). Although exoskeletons have a decisive function in the restoration of motor functions and can provide effective assistance for those that cannot afford human caregivers, access to these technologies is very limited. Usually, access to these technologies is to or through hospitals or rehabilitation centers where these expensive devices are financed through institutional means, such as insurance or subsidized health care (Fosch-Villaronga, Čartolovni & Pierce, 2020). Moreover, robotic engineers develop robotic devices primarily on the premise that the product should fit as many end-users as possible, ultimately serving the majority of users but failing to address the needs of a broader spectrum of users, including those that deviate from the majority of users. This is illustrated in the overwhelming overfocus on stroke rehabilitation as the only/primary object of study for rehabilitation.

In the field of exoskeleton development, children and their needs thus appear to be substantially overlooked, with some few exceptions (e.g., LokomatPro from Hocoma, the Atlas 2020, and 2030 from MarsiBionics, Trexo from Tréxō Robotics, and Strydr™ from Agilik technologies). The reason for this paucity of pediatric products can be attributed to 1) a lack of market appeal because of the investment or costs associated with product development for such a complex target population, 2) design challenges for pediatric exoskeletons in comparison to the design for adults, and 3) implementation challenges related to trust (Fosch-Villaronga, Čartolovni & Pierce, 2020). Nevertheless, just as measures are taken to incentivize and encourage the development of health-enhancing pharmaceuticals for all persons who may need them, the need to operationalize this imperative for pediatric exoskeletons in children's service is increasingly recognized and compelling.

*2.2.2.3. Developments in physical characteristics of Physically Assistive Robotics*

PARs are generally characterized by the variety of user interfaces and control systems that they incorporate and user perspectives that they generate. While from the device perspective, the operation of physically assistive robotic devices corresponds to the different control modes which they embody and that can vary from shared to supervisory controlled, from the user perspective, the operation of such devices may depend on the level of autonomy it requires (Arrichiello, et al., 2017). A shared-controlled mode involves the user controlling the system by continuously generating high-frequency motion commands and translating those commands from the control software into low-level functions. Through the supervisory-controlled mode, the user provides high-level low-frequency commands (e.g., to start/stop actions) while the system operates completely autonomously. In both cases, the control software must generate motion directives that realize the required action while considering safety, comfort, and efficiency (Arrichiello, et al., 2017).

As developments in assistive robotics advance, more dexterous and capable machines are believed to revolutionize how PCN-impaired individuals can interact within society and their surroundings and care for themselves more independently. While these machines become more capable, however, they often also become more complex. In answering how to control this added complexity, a confounding factor is that the more severe a person's motor impairment, the more limited the control interfaces available to them are to operate (Herlant, 2018). The control signals issued by these interfaces are lower in dimensionality and bandwidth. Thus, paradoxically, a greater need for sophisticated assistive devices is paired with a diminishing ability to control their additional complexity (Herlant, 2018).

User interfaces and control systems comprise the range of software and hardware components that allow a person with a disability to interact with their physically assistive robotic device (Brose et al., 2010). The robotic devices' operation modes are strictly connected to the Human-Machine Interface (HMI) used to generate and communicate commands (Arrichiello et al., 2017). Traditional interfaces often cover only a portion of the control space of more complex devices. PARs' ease of use depends on a workable user interface. While physical control mechanisms such as joysticks are among the simplest systems, the control's directional nature allows for natural device manipulation. The advent of three-dimensional joysticks has made control of devices with more degrees of freedom possible (Gieschke, Richter, Joos, et al., 2008). These mechanisms have been extended to include chin- and head interfaces, sip-and-puff (Fehr et al., 2000), voice control (Cagigas and Abascal, 2004), eye gaze direction (Yanco, 1998), EMG (Han et al., 2003), gesture- and intention-based human-robot interfaces, muscle-based robot interfaces, and Brain-Computer-Interfaces (BCIs). Among the different HMIs, BCIs represent a relatively new technology that has recently been proposed to drive wheelchairs (Bi, Fan, & Liu, 2013; Carlson & Millan, 2013) guide robots for telepresence (Leeb et al., 2015; Escolano, Antelis, & Minguez, 2012), control exoskeletons (Frisoli et al., 2012) and mobile robots (Gandhi et al., 2014; Riechmann, Finke, & Ritter, 2016). It has recently attracted large attention because it may be used in the absence of motion capability of the user with applications in different areas of assistive technologies such as motor recovery, entertainment, communication, and control (Arrichiello, Di Lillo, Di Vito, Antonelli, & Chiaverini, 2017). As control systems (continue to) become increasingly sophisticated and allow for larger "bandwidths" of information to be transferred from human to machine, increasingly sophisticated devices will be developed.

*2.2.2.4. Tensions between the desires of the patients and the research interests of the industry*

The shift from static automatic feeding robots to more blended multipurpose robotic arm applications such as iARM reveals a tension between the desires of the patients and the industry's research interests beyond the field of robot manipulators seeps into the domain of PAR as a whole. While the benefits of wheelchair-mounted robotic manipulators, among other performance-based assistive robots, include heightened autonomy and more control. The development of robotic assistance is often orchestrated with the needs of caregivers instead of care receivers in mind, with user autonomy being pitched as a problem to be solved rather than a faculty to be developed. The replacement of human assistance by robotic assistance to realize ADL, such as bathing or personal grooming (Mois & Beer, 2020), goes against the users' preferences to delegate assistive tasks. Notably, older adults prefer robot assistance for chores and tasks such as finding and fetching objects or housework (Smarr et al., 2014). In addition to the type of tasks assistance is being provided for, Brose et al. (2010) highlight how aesthetics are an essential consideration for device acceptance, namely that the appearance of the robot needs to be adapted to the task at hand (Broadbent et al., 2012). The question of device acceptance is essential when considering the growing ethical implications of the robotization of care. PARs such as *lifting robots* could potentially reinforce the notion that care can be reduced to caring for bodies (Parks, 2010) and impacting their dignity and lessening their self-respect (Sharkey, 2014).

# 3. Healthcare Service Robots

In addition to the robots used directly during medical procedures and other therapeutic applications, some robots facilitate care delivery and support doctors and other medical staff's work in another way. These robots help deliver medication and supplies, enhance patient-doctor contact, and clean hospital facilities (Cepolina & Muscolo, 2014). Although these robots may not fit into the classic picture of a healthcare robot, they still perform vital tasks within this sector and have distinct characteristics. These robots are called healthcare service robots (HSR).

While there is no commonly accepted definition for *service robots*, they are distinctive from industrial robots. In 1993, the Fraunhofer Institute for Manufacturing Engineering and Automation (Fraunhofer IPA) defined *service robots* as 'freely programmable kinematic devices that perform services semi-or fully automatically.' Within this context, *services* were 'tasks that do not contribute to the industrial manufacturing of goods but are the execution of useful work for humans and equipment' (Schraft, 1993). Over time, many definitions have been proposed, one of which by ISO. The ISO defines a *service robot* as 'a robot that performs useful tasks for humans or equipment, excluding industrial automation applications' (ISO, 2012).

In contrast, the International Federation of Robotics (IFR) emphasizes the robot's autonomy in their definition by defining the term as 'technical devices that perform tasks useful to humans' well-being in a semi or fully autonomous way' (IFR, 2015a). In healthcare, a *service robot* is 'any machinery in a clinical setting that can perform tasks, either partially or fully autonomously, to provide a useful service for healthcare delivery, including internal management', e.g., delivering and transporting goods or cleaning floors (Garmann-Johnsen, Mettler & Sprenger, 2014; IFR, 2014; Mettler, Sprenger, & Winter, 2017). As the task of defining service robots has continued to evolve, its definition has become more blurred, primarily due to the crossover between industry and service sectors. Contrary to industrial robots, which often operate in controlled domains or domains that are hostile to humans, service robots commonly function alongside humans and in a fairly uncontrolled environment (Mettler, Sprenger & Winter, 2017). An example of this can be found in mobile robots and Automated Guided Vehicles (AGV) used in industrial automation applications and as service robots in new environments such as hospitals (Holland et al., 2021).

The story of HSRs has not always been successful (Stone et al., 2016). Already more than thirty years ago, a service robot was introduced to function as a courier robot in hospitals. The robot, called HelpMate, was supposed to carry around deliveries such as meals and medical records (Evans et al., 1989). This type of robot's development was relatively simple and similar to the other robot developments in the healthcare sector, but their use has not been mainstream. With the high and rising costs in the healthcare sector, social pressure for lower costs, labor shortages, and an increasingly sick and aging population, the increasing investments in robots in the healthcare sector have proved very promising (Simshaw et al., 2015). It is predicted that the demand for professional service robots to support healthcare staff will reach 38 billion USD by 2022 (Müller, 2018) to lower the workload of healthcare staff and aid in complex tasks that need to be carried out (Taylor et al., 2016).

HSRs make the delivery of care and hospital management more effective and quick, reduce labor costs for repetitive and often tedious tasks, and improve healthcare practitioners' work (Fosch-Villaronga, 2019). HSRs can, therefore, replace jobs or assist in performing tasks in the healthcare setting. Although heavily automated, however, many tasks performed by HSRs still require some degree of human intervention. For example, delivering food with an unmanned ground vehicle would make it possible to transport food to a specific hospital room, requiring a nurse to give the food to the patient. As pointed out earlier within the context of SARs, developments in HSRs cannot be aligned or limited to a single purpose, either. The field of HSRs is not oriented towards a single issue but instead covers a wide field of application of these robots, which is not only limited to the medical field. For example, robots such as the delivery robot Relay is used in hotels, hospitals, and public spaces, and the VGo telepresence robots are used in healthcare, education, and business. The possible non-medical purposes of such robots are essential to consider given that healthcare is a sensitive context of use, often requiring additional safety layers.

### 3.1. Healthcare Service Robots ecosystem

HSRs are believed to streamline routine tasks, reduce the physical demands on human workers, and ensure more consistent processes (Mettler, Sprenger, & Winter, 2017). These robots can keep track of inventory and place timely orders, helping make sure supplies, equipment, and medication are where they are needed at the relevant time. Mobile cleaning and disinfection robots allow hospital rooms to be sanitized and readied for incoming patients quickly. HSRs can also be an excellent tool for sanitary reasons, which are vital in care settings. Although advances in robotic technology have traditionally been in the manufacturing industry due to the need for collaborative robots, this is no longer the case in the service sectors, especially in the healthcare sector. The lack of emphasis on the healthcare sector has led to new opportunities in developing service robots that aid patients with illnesses, cognition challenges, and disabilities. Also, the COVID-19 pandemic has further triggered the development of service robots in the healthcare sector to overcome the difficulties and hardships caused by this virus (Aymerich-Franch & Ferrer, 2020). Clinical care was the second largest set of uses (Murphy, Gandudi & Adams, 2020) for robotic devices throughout the pandemic, all with ground robots. This category represents applications related to the diagnosis and acute healthcare of patients with the coronavirus. More specifically, the use of robots protected healthcare workers by enabling them to work remotely, allowed them to delegate unskilled tasks such as meal delivery and disinfection, and to cope with surges in demand; the largest reported use for robots in clinical care throughout the COVID-19 pandemic are healthcare telepresence, disinfecting hospitals or clinic, prescription and meal dispensing, use of telepresence robots in processing the intake of patients and handling families, essentially protecting the receptionists and clerks, in enabling families to visit patients remotely, and in automating inventory management for a hospital floor (Murphy, Gandudi & Adams, 2020).

A clear example of this can be seen in the surgical environment (Zemmar, Lozano, & Nelson, 2020). Robots have long served humans to protect them from hazardous tasks. The pandemic has brought about several threats and restrictions to our society in general and the healthcare sector in particular. As a result, various potential implementations have been proposed to put robots to use in healthcare and beyond to face these challenges (Zemmar, Lozano, & Nelson, 2020). Besides surgical robots, HSRs offer an attractive tool for use in activities such as disinfecting streets, measuring patients' temperature at a distance, and assisting medical staff in providing a safe and remote communication instrument within a care setting. HSRs thus ensure hygiene, limit the direct contact between doctors, nurses, and patients to reduce the risk of infection, and at the same time, reduce the number of protective masks and gowns needed for everyone, making the need for HSR far greater than before (WEF, 2020; Zemmar, Lozano, & Nelson, 2020). Service robots are advantageous because they prevent the spread of infection, reduce human error, and allow front-line staff to minimize direct contact, focusing their attention on higher priority tasks and creating separation from direct exposure to infection (Holland et al., 2021).

Although HSRs are broadly defined, a wide variety of HSRs exists. The literature related to HSRs lacks specific categories, with most HSRs merely being described based on their characteristics. Nevertheless, a remarkable similarity between all HSRs is that they have been developed to make the hospital's daily processes more manageable and efficient. Recently, the industry has shown a growing interest in developing robots to assist nurses in hospitals and clinics, with the robotic nursing assistants functioning under the nurses' direct control. Such robotic nurses are designed to act as a teammate, helping in the performance of non-critical tasks (e.g., fetching supplies, and giving nurses more time to focus on critical tasks, such as caring for patients).

Especially throughout the COVID-19 pandemic, it became clear how vulnerable nurses are due to shortages of Personal Protective Equipment (PPE) (Ranney, Griffeth & Jha, 2020). Even among the best-equipped centers across the globe and the most developed nations, a shortage of protective equipment has become a key concern (Zemmar, Lozano, & Nelson, 2020). Another factor that is fundamentally limiting in this context is the healthcare labor capacity. While machines in the healthcare environment could operate beyond maximum capacity over more extended periods, human healthcare workers cannot follow this pace, which is incredibly limiting during high-demand periods (Zemmar, Lozano, & Nelson, 2020). The adoption and implementation of robots in the healthcare environment can reduce direct contact between the healthcare provider and patient and serve at maximum capacity under extraordinary and high-demand circumstances. This realization has acted as a catalyst for further developing service robots in the healthcare setting, as robots, unlike human nurses and hospital staff, are not vulnerable to viruses or other microorganisms. In this respect, it seems they are of significant relevance during pandemics.

## 3.2. Categories

Within the context of HSRs, we distinguish between a wide variety of categories and types. We can differentiate between those HSRs that completely take over medical staff tasks, HSRs that support routine (non-medical) tasks, and those that facilitate specific tasks. Another critical characteristic of HSRs relates to their level of autonomy. Some HSRs, such as the delivery robot TUG, function autonomously, while others require some degree of human intervention. Also, some HSRs function autonomously to a certain extent and assist, rather than replace, medical staff. With this type of HSRs, several stakeholders (e.g., medical professionals and patients) are involved in the robot's use and functioning to realize its purpose. For example, the InTouch Health telepresence robots provide virtual care and make it possible for a medical professional to contact patients from a distance. However, without the stakeholders' involvement, the robot would not fulfill its envisioned function.

The differences, overarching similarities, and trends found when comparing existing HSRs reveal different HSR categories. These are routine task robots, telepresence robots, disinfectant robots (and types within these categories), delivery robots, automated dispensing robots, remote inpatient care robots, remote outpatient care robots, infection prevention robots, and general cleaning robots (*see* Table 7). At the heart of this distinction lies the specificity of assisting or replacing medical staff, the robot's autonomy level, and their primary function.

| Classification of Healthcare Service Robots | |
|---|---|
| **Category** | **Subcategory** |
| Routine task robots | - Delivery<br>    - Automated Guided Vehicles<br>    - Serving robots<br>    - Mobile robots or platform<br>    - Drones<br>- Automated dispensing<br>- Healthcare administration |
| Telepresence robots | - Remote Inpatient Care (RIC)<br>- Remote Outpatient Care (ROC) |
| Disinfectant robots | - Infection prevention<br>- General cleaning |

**Table 7**. Classification of Healthcare Service Robots

### 3.2.1. Routine task robots

Routine task robots are autonomous and mobile robots that assist medical staff with daily routine tasks such as delivering food and medicine, carrying linens, pushing beds, or transferring lab specimens. In some cases, these robots even replace the medical staff as a whole. Although these robots are fully autonomous and mobile, they are not necessarily anthropomorphic or social (Simshaw et al., 2015). Routine task robots are designed to perform everyday tasks to relieve the pressure on the medical staff. Three types of routine task robots can be distinguished within the category of HSRs, namely 1) routine task robots designed to deliver medical goods and to move around appliances in hospitals or other medical environments (delivery robots); 2) routine task robots used in automated processes, such as medicine dispensing in (hospital) pharmacies (automated dispensing robots), and 3) administrative robots used in healthcare management.

*Delivery robots*

An important and often underestimated aspect of healthcare is its underlying logistics. Especially in a hospital, it is vitally important for the logistic system to organize and maintain its material flow. As many materials are transported in hospitals each day, such as medicine, medical supplies, laboratory samples, food, and linen, the healthcare sector is ideal for this application, especially in logistics. For example, nurses spend as much as 30% of their time away from patient bedside care, tracking down medications, supplies, and lab results, as well as carrying out logistic duties (Holland et al., 2021). Medical staff is also often directly exposed to the health risks present in the hospital setting. As a result, an emerging application for autonomous navigating robots is hospital delivery. Due to the need to reduce these logistic processes' operational costs and deal more efficiently with the increasing pressure on the supporting logistic functions and the rising demand for materials and equipment, a growing interest in logistics automation in hospitals has been identified. By automating the logistics within the hospital setting, a more flexible transportation plan can be followed irrespective of staff availability, and which can be planned during night and day. Moreover, transportation routes can be optimized, and hospitals' effectiveness can be improved, thereby increasing the quality of care (Mettler, Sprenger, & Winter, 2017).

Another type of delivery robot currently used in the medical sphere is serving robots. These robots carry out heavy-duty tasks in hospitals where pushing and pulling of material is required and are also deployed to supply food and beverages, dispense drugs, remove unclean laundry, deliver fresh bed linen, and transport regular and contaminated waste to/from various patients residing in hospital (Ozkil et al., 2009; Mettler, Sprenger & Winter, 2017). Examples of such robots include the Panasonic Autonomous Delivery Robots - HOSPI, the TUG autonomous service robot, RELAY robot, and LoRobot L1. Since the 1950s, AGVs have optimized logistics in factories, hospitals, and homes (Vis, 2006). AGV's are capable of transporting materials through wire guidance, inertial guidance, or laser guidance, which means that they depend to a certain extent on a predefined route and system. For example, the TransCar© robot is a self-guided delivery robot that can deliver medication, linens, and meals in the hospital environment. More recently, mobile robots are rising (Acosta Calderon, Mohan & Ng, 2015). Within the healthcare setting, HelpMate was the first mobile robot designed and developed to deliver pharmacy supplies and patient records between hospital departments and nursing stations (Evans, 1994). The robot used sensor-based motion planning algorithms that were revolutionary when they addressed navigating in unknown and unstructured environments (Evans, 1994). The main difference between these mobile robots or robotic platforms and AGVs is that instead of using an infrastructure, the robot plans and senses routes by itself. In other words, these mobile robots have a higher level of autonomy. Another example of this type of delivery robot is the I-Merc robot, a mobile robot specializing in delivering meals in hospitals. Besides functioning almost entirely autonomously, this robot can also pay special attention to personalized diet information (Canas, Silva & Cardeira, 2006).

Another type of delivery robot currently used in the medical sphere is drones, which have traditionally been used outside the healthcare environment. Nevertheless, they are increasingly used as innovative tools for medical equipment delivery: medicine, defibrillators, blood samples, and vaccines. For example, autonomous drones use Global Positioning Systems (GPS) and other sensors to navigate automated ground stations to deliver medications in remote locations that lack adequate roads. Drone technology and its components, such as GPS and lithium batteries, are available and improving at a rapid pace (Scott & Scott, 2017). The timely delivery of relevant goods in healthcare settings is vitally important. The use of healthcare delivery drones makes it possible to quickly and efficiently deliver medical care to places where this would not usually be the case. Drones enable developing countries to leapfrog ahead with healthcare delivery to remote locations, even with unreliable road infrastructure. For example, Matternet drones were used to deliver medicines after Haiti's earthquake in 2010, and in Germany, the Parcelcopter - designed by DHL Parcel - provides medications, materials, and blood samples (Scott & Scott, 2017).

*Automated dispensing robots*

In the hospital and home, the pharmacy acts as an extension of specialists' medical care in hospitals and fulfills an essential role within the healthcare sector. For nearly everyone, taking the right medicine and the right amount is vital. Due to the complexities within the medication-use process, errors are inevitable. In the final decades of the 20th century, automated medication dispensing systems started being introduced and implemented in (hospital) pharmacies. Their purpose is to minimize medication dispensing errors, save time, and secure the drug and administration process (van den Bempt et al., 2009; Boyd & Chaffee, 2019). Since their introduction, several studies have shown a moderate decrease in medication dispensing errors and time (Jones, Crane & Trussel, 1989). Besides preventing medication errors, the use of robots and automated processes within the medication-use process also reduces costs. For example, the German company BD Rowa developed several dispensing robots that are now used in pharmacies and hospitals worldwide to provide faster and better care. The robots replace human employees and are entirely operated autonomously.

Moreover, throughout the COVID-19 pandemic, automated dispensing robots have gained significant interest within the healthcare sector. The third-largest use of robots throughout this pandemic has been identified within the field of prescription and meal dispensing, whereby carts are navigated autonomously through a hospital. As with disinfection robots, these types of robots were already commercially available before the pandemic, but their visibility and use increased as a means of coping with the surge in patients and the need to free the healthcare workers to spend more time on direct and compassionate care (Murphy, Gandudi & Adams, 2020).

*Healthcare administration*

Within the context of healthcare administration, robots are increasingly being deployed to streamline better routine administrative processes in hospitals and other clinical environments. These robots are preferably used at a hospital's reception to disseminate information about various units/sections of the hospital and guide patients and visitors. Examples of such robots include Pepper and Dinsow 4 robot. These robots can handle several visitors without becoming tired and direct them to the physician of their choice. They are exceptionally well received by children coming to the hospital, who experience their visit to the hospital as more pleasurable due to the interaction with the robot (Khan, Siddique & Lee, 2020). As they are right now, though, the field seems to point more at using artificial intelligence systems to boost healthcare administration and not that much fully embodied robot technology.

### 3.2.2. Telepresence robots

Telepresence refers to a set of technologies used to create a sense of physical presence at a remote place. Telepresence robots allow human operators to be virtually present and interact remotely through robot mobility and bidirectional live audio and video feeds (Koceski & Koceska, 2016). Teleoperated from a distant location, a mobile robot with some autonomous capabilities can become a particularly beneficial telehealth application tool. Recently, there is a growing interest in developing telepresence robot systems for older adults' wellbeing (Koceski & Koceska, 2016). Assistive technologies for telementoring in homes constitute a very promising avenue to decrease the load on the health care system, reduce hospitalization periods, and improve quality of life (Michaud et al., 2007). Like other HSRs, telepresence robots' functions are not limited to healthcare but widely cover business and educational environments for videoconferencing or other commercial activities too. Telepresence robots used in healthcare proactively socially engage with users, creating an interaction with the person to give assistance and support in certain ADL and care (Broekens, Heerink & Rosendal, 2009; Feil-Seifer, & Mataric, 2005). Also, these robots can collect medical data about the vital signs of patients that are important for doctors and caregivers. They can be used for social interaction with other people (e.g., family members, friends, doctors, or caregivers). They can also help the elderly overcome a sense of social isolation and loneliness, affecting older people's physical, mental, and emotional health (Moren-Cross & Lin, 2006; Søraa et al., 2021).

Telepresence robots' ultimate goal in healthcare is to provide specialized healthcare services over long distances, thereby bridging the physical gap between the medical professional and the patient. Telepresence robots make it possible to bring specialists and experts to remote areas where these services are currently unavailable (Kritzler, Murr, & Michahelles, 2016). They are also expected to play an essential role in the healthcare provisions in third-world countries and war zones (Avgousti et al., 2016). As such, telepresence robots are believed to ensure a better quality of life in underdeveloped, isolated, or remote areas. They also reduce the risks of transmitting infectious diseases between humans, which has become particularly important since the COVID-19 pandemic outbreak (Aymerich-Franch & Ferrer, 2020). Even more so, within this context, the most prominent reported use of robotics within clinical care during the COVID-19 pandemic was for healthcare telepresence, including the use of teleoperation by doctors and nurses to interact with patients for diagnosis and treatment. These robots were either commercially available or adapted from commercially available bases (Murphy, Gandudi & Adams, 2020). Moreover, telepresence robots lower healthcare-related costs and inconvenience and provide better access to crucial healthcare-related information. Nevertheless, delivering medical information via robots raises questions about which type of information is appropriate for this channel, whether it includes medical decisions and diagnosis or if it should be reserved solely for information purposes.

Generally, two types of telepresence robots can be distinguished within the context of the HSR category, based on the user, the remote environment where they function, and the stakeholders that interact with the robot, namely Remote Inpatient Care (RIC) and Remote Outpatient Care (ROC). Concerning RIC, the doctor is the primary user, using the robot within the hospital environment and interacting with nurses and patients. This type of telepresence robot is mainly used for consultations and check-ups during hospitalizations. One of the first telepresence robot systems allowing care assistance for the elderly was the Physician-Robot system, developed as a result of the InTouch Health Company and Johns Hopkins University's cooperation. This system enables physicians to visit their hospitalized patients more frequently (Koceski & Koceska, 2016). More recently, InTouchHealth has developed the RP-VITA platform, a successor of the RP-7 that combines autonomous navigation and mobility, allowing doctors to monitor patients remotely (InTouch, 2011).

Regarding ROC telepresence robots, while doctors remain the primary user, the remote environment in which the robot functions is the non-clinical environment (e.g., the patient's home). Besides the doctors and patients, the robot also interacts with the caregivers that surround the patients. Examples of ROC telepresence robots are the VGo robots by VGo Communications (Tsui & Yanco, 2013). These robots are not different in terms of function and construction but rather in how they are implemented within the healthcare sector. From this follows that some telepresence robots are both used as RIC as well as ROC robots.

### 3.2.3. Disinfectant robots

While not new to the healthcare environment, the COVID-19 pandemic has led to a surge in the development and adoption of disinfectant robots in the healthcare domain. Due to the severe influx of patients and shortage of medical staff, and in an effort to reduce the exposure of medical staff to patients, and also to maintain the social distancing guidelines, robots were deployed in hospitals and field hospitals to assist, among other things, to clean and sterilize (disinfect) (Gupta et al., 2021). To get one step ahead in terms of safety, such robots allow healthcare workers on different levels to remotely monitor and manage their daily operations by robotic and autonomous solutions so that the risk of exposure should remain low. Moreover, using robotics and autonomous solutions restricts human interference to fewer subject areas. Hence the risk of exposure is low (Gupta et al., 2021). Generally, disinfectant robots can be divided into two subcategories, namely 1) disinfectant used for general cleaning of hospital environments and other in- and outdoor spaces; and 2) disinfectant robots used to sterilize work surfaces and medical equipment to prevent Healthcare-Associated Infections (HAI) from infecting patients (Khan, Siddique & Lee, 2020).

*Infection prevention*

It is a long and well-known fact that the danger of new bacteria and infection by pathogens in healthcare environments, especially hospitals, is serious (Begić, 2017). The consequences of HAIs include considerable pain, suffering, and even death (Begić, 2017), thereby constituting major problems and high costs for the modern health sector. As cleaning and disinfection are expensive and not sufficiently effective due to inaccessible areas, advancements in robotic technology have triggered the usability of robotics in the next generation healthcare system (Kaiser et al., 2021), among which to disinfect hospitals and other healthcare environments. With sterilization methods not always being readily available and accessible, these robots offer a cost-effective solution to the manual disinfection of surfaces and objects, both in terms of time and minimization of the risk of exposure, to meet the remaining need for surfaces and objects to be disinfected (Kovack et al., 2017).

Robots are believed to be of significant value in reducing the risk of hospital infections, which can be transmitted in many ways (e.g., remote control, door handles or cabinets, a button to call for help). As a result, many new disinfection robots have been developed to help clean and disinfect these high-risk areas, including human support robots to sanitize high contact points and automated solutions for cleaning walls and floors. Sterilization robots are designed to emit a specific wavelength of ultraviolet light to the exposed surface to kill viruses and bacteria without exposing human personnel to infection (Guridi et al., 2019). Mainly due to the congested and complicated workspace in which they operate, disinfectant robots are generally remotely controlled from a safe distance. Many hospitals worldwide are already accepting these robots on a trial and real-time basis to get rid of viruses and bacteria in rooms and halls, and on door handles, especially since the outbreak of the COVID-19 pandemic. Even more so, this pandemic has accelerated the testing of robots and drones in public use, as officials seek out the most expedient and safe way to grapple with the outbreak and limit contamination and spread of the virus (Gupta et al., 2021). An example of such a robot is the UV-C disinfection robot, which provides an economical and effective measure in limiting bacteria's spread by exposing the bacteria to UV-C.

HAIs impose significant threats to vulnerable patients in hospitals. These infections are infections that the patient can potentially contract while being treated for something else and annually cause thousands of deaths and huge healthcare costs. One example of an HAI is the Surgical Site Infection (SSI), which is an infection in the part of the body where previously the surgery took place. Because of the high risks of HAIs, hospitals use strict prevention protocols to prevent these infections. These protocols mainly focus on the disinfection of hospital equipment, vehicles, and areas. The use of disinfection robots by hospitals to comply with these protocols is widespread. Especially in the light of the recent COVID-19 pandemic, there is a great demand and decrease of such robots.

Within this context, the second-largest reported use was for rapidly disinfecting the hospital or clinic. The majority of robots use UVC light to perform gross disinfection, followed by a human wiping down surfaces likely to be missed by the robot. Several models of this type of robot were already commercially available and used to prevent hospital-acquired infections before the outbreak of the pandemic (Murphy, Gandudi & Adams, 2020). For disinfection in the healthcare environment, autonomous bots are often being transformed or adjusted to reassign tasks. For instance, collaborative robots for machine tending and warehouse rack-stacking are being deployed in the fight against novel coronavirus. Drones originally designed to spray pesticides for agricultural applications have been quickly repurposed to spray disinfectants to fight against COVID-19 (Kaiser et al., 2021). An example of this is the Chinese robot company, Youibot, which has rolled out a multi-purpose robot that can monitor customer's temperatures using infrared cameras during the daytime and disinfect surfaces with the help of ultraviolet (UV) light in high-traffic areas, including hospitals, at night. Other examples include the Connor UVC Disinfection Robot, DJI, which is one of the companies that shared the responsibility to disinfect millions of square meters in China, and HAI by Xenex, which is widely adopted worldwide.

*General cleaning*

Robots for general cleaning purposes are on the rise since the outbreak of the COVID-19 pandemic. In China, for example, robots have been assigned multiple tasks to minimize the spread of COVID-19, such as utilizing them for cleaning and food preparation jobs in infected areas hazardous for humans (Khan, Siddique & Lee, 2020). Generally, robots are used in hospital cleaning, using dry vacuum and mopping (Prassler et al., 2000). Such robots are an integral part of disinfecting hospitals to remove germs and pesticides. Examples of such robots include the Roomba cleaning robot by iRobot, an intelligent navigating vacuum pump for dry/wet mopping, UVD robot by UVD Robots ApS, an UV radiation-based device used to disinfect hospital premises from microbes. The Peanut robot used to clean washrooms of hospitals by using a highly dynamic robotic gripper and sensing system, and Swingobot 2000 by TASKI, a heavy-duty cleaning robot for cleaning hospital floors autonomously (Khan, Siddique & Lee, 2020). The research in this area seems fast-developing with increased attention during the COVID-19 pandemic.

## 3.3. Considerations for healthcare robot policymaking

### 3.3.1. Autonomy and accessibility considerations for HSRs

Healthcare robotics span a wide range of robots used for different use cases and varying autonomy levels. Today's most prominent use in clinical environments is stationary or teleoperated devices, such as robots that assist with surgery, rehabilitation, information, or drug-dispensing (Ahn et al., 2014; Barrett et al., 2012; Bepko Jr et al., 2009; Berlinger, 2006). Inherent to such robots, actions and tasks require different degrees of human intervention, ranging from none to complete or fixed installation to function appropriately (Mettler, Sprenger & Winter, 2017). The autonomy levels for HSRs and the subsequent interplay with the role and responsibilities of the humans involved in HSRs have not been defined yet.

While there is evidence that these applications are helpful, most people would not directly associate robots with them (Diprose et al., 2012). Mainly when focussing on HSRs, this becomes a topic requiring special attention from policymakers. Under the label service robots, these robots are designed to '[…] operate semi- or fully autonomously to perform services useful to the well-being of humans […]' (IFR, 2016). This definition considerably expands the digitization of work processes in hospitals and other health organizations (Yoo et al., 2010). The HSRs studied for this research include robots with varying degrees of autonomy, including those with fully autonomous capacities and those which do not require a controlled environment or fixed installation in clinical settings to deliver intangible, automated, or personalized services to humans.

For instance, one could imagine a routine task logistics robot that autonomously navigates in an uncontrolled environment (e.g., crowded hallways or areas with many obstacles) to deliver meal trays, sterile supplies, or blood samples, mail (Deery, 1997; Kirschling et al., 2009). Although many such HSRs have already been designed to accommodate such deliveries, they still appear not to offer the most flexible solutions as they are not easily modified (Jeon & Lee, 2017). They are often costly and present issues regarding installation in older buildings, navigating through narrow, snake-like corridors, and high human footfall (Holland et al., 2021). An example of this relates to robots used for logistics in hospitals. While these robots help the transport of medical supplies, they are limited by their speed and the size of the hospitals. Many of these robots are only suitable for corridors and not in more complex environments such as areas with patients or surgical wards (Holland et al., 2021). As a result, human intervention may still be needed to perform such tasks initially assigned to the robot. Although it is becoming increasingly clear that the deployment of HSRs significantly increases overall efficiency in the healthcare sector while simultaneously protecting front-line hospital staff from the risk of disease, eliminating human error, and allowing staff to focus their attention on higher priority tasks, health policies should consider a move towards lower-cost sensor alternatives to make these robots more accessible. Moreover, further research into machine learning is needed before these types of service robots can be developed further (Holland et al., 2021).

### 3.3.2. The need for an extra layer of safety for HSRs

There are both advantages and challenges associated with robots in disease prevention and management (sterilization and disinfection), logistics, telehealth, and social care. Service robots used for disease prevention and control are highly beneficial to society and public health because they can work around the clock to disinfect areas, thus relieving healthcare staff and reducing person-to-person contact. However, this domain poses challenges. When defining the workspace for HSRs, it is crucial to consider the human factor since the workspace between humans and robots is continuously shrinking. While HSRs can be significantly beneficial to the healthcare sector, this development, in turn, raises novel occupational safety and health issues. It is also unclear whether existing legal frameworks and guidelines entail sufficient adequate responses to health safety-related issues caused by the deployment of HSRs, which are increasingly autonomous, in uncontrolled environments. A clear example of this problem lies in the deployment of drones in the healthcare context. Drones can provide beneficial and humanitarian applications, especially healthcare, making the delivery to inaccessible locations more ubiquitous soon.

Nevertheless, these drones are often criticized for inadequate regulation, safety issues, security, and privacy abuse (Gupta et al., 2021). Current safety standards for industrial robots and robotic divides (e.g., ISO 10218-1:2011) are established to ensure that robots are not hazardous to humans while performing their tasks. Within this context, the safety of humans is thus paramount in the deployment of robots. Another often overlooked yet essential safety measure to ensure robotics operations' safety is privacy and (cyber)security. The increasing connectivity of robots to the internet, such as ethernet or Wi-Fi and soon via 5G, introduces significant security vulnerabilities. Since robot manufacturers are not well-versed with cybersecurity standards as other industries are, since they are not seen as potential targets for cyber attacks (Gupta et al., 2021; Clark, Doran & Andel, 2017; Fosch-Villaronga & Mahler, 2021). In this sense, policies and (additional) safeguards ought to be put in place to ensure that robot manufacturers follow sufficiently adequate legal frameworks as other industries do. The same applies to the issue with the privacy of users' data concerning telepresence robots and personal assistant robots, indicating the need for a strict protocol with regards to where the data are stored, who can access this data, and whether the information being recorded is done so with consent (Barabas et al., 2015).

### 3.3.3. HSRs and the impact of COVID-19 pandemic

The deployment of robots during the COVID-19 pandemic is happening quicker than expected (Gupta et al., 2021). While technologies like telemedicine, telepresence, autonomous delivery robots, and sterilization robots have shown significant pragmatic promise well before the COVID-19 pandemic, they did not achieve the success everyone hoped for various reasons (Gupta et al., 2021). Nevertheless, in the current COVID-19 world, many of these technologies have been fast-tracked to provide a near-to-normal lifestyle and protect citizens while simultaneously protecting healthcare workers from exposure to a contaminated environment and patient and alleviating the pressure on the healthcare sector worldwide. As such, telehealth technologies have helped in avoiding the collapsible state of the health care sector while at the same time providing a safe, cost-effective, accessible, and convenient alternative in real-time. Futuristic seeming services like food and essential delivery via robots, robot nurses in hospitals, and disinfestation robots might have taken a few more years to become the norm and are now rapidly being put to use on a global scale.

Beyond the clinical context, HSR's - especially telepresence technologies - have increasingly been deployed in another effort to reduce human contact and curb the spread of the virus. These technologies include wearable devices and digital contact tracing apps used in various countries to identify individuals who have been in contact with an infected individual. These apps use Bluetooth and the user's geographical location, which is obtained via either the cellular network or an app installed on a smartphone to properly function, sparking concerns about mass surveillance, raising questions on privacy and data protection, and presenting regulatory challenges. The COVID-19 pandemic has allowed the deployment of these technologies to become better, cheaper, and more accessible. In this respect, policymakers will have to catch up and take regulatory action sooner rather than later to address the (adverse) implications of the rapid adoption of these technologies and the compromises on user safety that have been made due to exceptional circumstances. For example, the pandemic may be the right inducement for regulatory bodies to call for an open-source platform for all applications requiring access to sensitive user data (Gupta et al., 2021).

Another example is the waiver policy applicable to drone operators, which allows them to operate drones freely. This, however, compromises safety. While the skill of the drone operator may well justify the trade-off between productivity and safety - which is especially necessary in times of crisis -, the human factor should nevertheless be considered when developing policies for the deployment of such robots, especially in the sensitive domain of healthcare (Gupta et al., 2021).

### 3.3.4. Safeguarding the human factor in the HSR domain

A significant challenge in the field of HSRs is their acceptance (Holland et al., 2021). Exposure to new technology alone does not automatically lead to a positive attitude towards robotic technology (Kristoffersson et al., 2011). In this sense, Holland and colleagues (2021) argue that it would be beneficial to hold public engagement campaigns to promote positive attitudes and acceptance towards service robots. The purpose would be to introduce and train healthcare staff on operating these robots while also giving service robots more exposure, making them visible in everyday environments. Furthermore, although robots are only accepted insofar as they benefit one's work, making it more efficient and pleasant, there is still the fear that robots will replace their human counterparts, resulting in job losses. Several studies indicate a strong need for staff education, training, and involvement (Garmann-Johnsen, Mettler & Sprenger, 2014). From an economic perspective, as HSRs are deployed at an increasing rate and on a large scale, we will have to account for human training, especially for collaborative robots, which requires more investment than robot-only tasks.

For this reason, the nature of the tasks assigned to the relevant robots should be considered to gauge the investment required to deploy robots on a commercial scale, thereby also influencing the acceptance of such robots. At the same time, staff must be prepared, and technology readiness must be established (Peronard 2013). In this sense, hospital staff's perception of workload and stress is a positive determinant for motivating the shift to robot couriers (Mendell et al. 1991) and triggering automation's imminent need. Therefore, the whole staff of the hospital must be adequately educated on how to deal with autonomous service robots (Kirschling et al. 2009), as has also been pointed out in the context of surgery robots. Moreover, while existing discussions of the adoption of service robots in healthcare mirror high hopes and aspirations and often reflect the opinions of policy-makers and technologists (Hagele, 2016), they nevertheless fail to account for health workers (Mettler, Sprenger & Winter, 2017). This calls into existence the need for the relevant technology developers and their users to collaborate, thereby taking into account primary users, secondary users, and bystanders (Huttenrauch et al. 2004).

### 3.3.5. Lack in clarity on possible use cases and users of HSRs

Research has pointed out the lack of clarity regarding the full potential of service robots in healthcare. It thus indicates the need for further knowledge regarding different use cases and the implications for the hospital organization when employing service robotics (Garmann-Johnsen, Mettler & Sprenger, 2014). While it is clear that, in general, there are many fields of application for service robots, it remains unclear in which cases and to what extent such devices would be supported in the hospital context (Garmann-Johnsen, Mettler & Sprenger, 2014). When introducing service robotics into the hospital, one has to keep in mind that the tasks and processes supported by the robot are not independent but are related to some extent to the existing hospital systems and processes (Garmann-Johnsen, Mettler & Sprenger, 2014). Many new automation opportunities are eminent in clinical support systems (Garmann-Johnsen, Mettler & Sprenger, 2014). Yet, it is unclear how much integration is needed to fully leverage the potential of service robotics, while at the same time, it is also highly complicated to keep complexity and financial considerations in mind. This thus remains a very complex topic that requires further clarification.

Moreover, it remains unclear who the actual users of HSRs are and how these service robots 'fit' into existing social systems (Garmann-Johnsen, Mettler & Sprenger, 2014). In the clinical context, not one uniform user can be identified. There are different kinds of hospitals, but even within one hospital, different users can be distinguished depending on the service robotics' use cases (Garmann-Johnsen, Mettler & Sprenger, 2014). Users can also be determined according to their attitudes towards service robotics. Different user groups hold different values that could act as drivers or inhibitors to introduce service robotics (Garmann-Johnsen, Mettler & Sprenger, 2014). While it is thus crucial that HSRs are compatible with existing processes and systems in the hospital context, it is equally important that these devices are consistent with the attitudes and values present in the hospital context.

# Conclusions, highlights & key take-aways for policymakers

Inserting robots into healthcare is not straightforward, and the field is not very well defined yet (Riek, 2017; Fosch-Villaronga, 2019a). Robot surgeons, physical/socially assistive, and Healthcare Service Robots largely differ in embodiment and context of use, often navigating between medical and non-medical device categories. Current frameworks also overfocus on physical safety, neglecting other essential aspects like security, privacy, discrimination, psychological aspects, and diversity, which play a crucial role in robot safety. As a result, healthcare robots fail to provide an adequate level of safety 'in the wild' (Gruber, 2019). Moreover, increased levels of autonomy and complex interaction with humans blur practitioners and developers' roles and responsibilities and affect society (Carr, 2011; Yang et al., 2017; Boucher et al., 2020).

Unfortunately, as the pace of technology dramatically accelerates, our understanding of its implications and its regulation does not keep up with this pace (Collingridge, 1980; Marchant, 2011; Newlands et al., 2020). On the contrary, current proposals have so far failed to frame healthcare robotic technology adequately, although it is a remarkably sensitive domain of application. Given this uncertainty, neither regulators nor developers know what needs to be done, while users' rights are already at stake (Fosch-Villaronga & Heldeweg, 2018). Our contribution aimed at filling in the existing gap and lack of clarity currently experienced within healthcare robotics and its governance.

To this end, we provided a structured overview of and further elaboration on the main categories now established, their intended purpose, use, and main characteristics, and complemented these findings with policy recommendations to help policymakers unravel an optimal regulatory framing for healthcare robot technologies. In this last chapter, we provide a summary of the main highlights from the different chapters, they key take-aways for policymakers and a final list of policy recommendations.

Future work will include a systematic literature review on the topic of healthcare robotics that includes a more comprehensive overview of the field and other types of robots, including nanorobotics, neuro-rehabilitation, and AI-powered devices.

## Chapter highlights

| On Healthcare Robots \| chapter highlights |
|---|
| **Introduction** |
| Robotics have increased productivity and resource efficiency in the industrial and retail sectors, and now there is an emerging interest in realizing a comparable transformation in healthcare. |
| Despite several clear benefits, systems that exert direct control over the world can cause harm in a way that humans cannot necessarily correct or oversee. |
| Although healthcare is a strikingly sensitive domain of application, it is still unclear whether and how healthcare robots are currently regulated or should be regulated. |
| Before regulating the field of healthcare robots, it is essential to map the major state of the art developments in healthcare robotics, their capabilities and applications, and the challenges we face as a result of their integration within the healthcare environment |
| **Chapter 1:** Defining healthcare robotics and their categories |
| Within the field of law, definitions play a pivotal role, functioning as a mechanism for avoiding ambiguity in interpretation, and most importantly, warranting the application of the law to a case. |
| While there is no generally accepted definition for the term robot, for this contribution, we define a robot as a movable machine that performs tasks either automatically or with a degree of autonomy. |
| The wide adoption of robot technologies within the field of healthcare lies 1) in the wish to reduce expenditure, 2) broaden access to healthcare and 3) in the desire to improve prevention and patient safety. |
| The EFMN defines healthcare robots as 'systems able to perform coordinated mechatronic actions (force or movement exertions) based on processing information acquired through sensor technology, to support the functioning of impaired individuals, medical interventions, care and rehabilitation of patients and also individuals in prevention programs.' |
| This definition raises questions on which legal categories (e.g., product, medical devices) healthcare robots are. |
| Within the field of healthcare robotics, three main categories of healthcare robots can be distinguished: surgical robots, assistive robots (spanning both physically assistive and socially assistive robots), and healthcare service robots. |
| **Chapter 2:** Defining the healthcare robot ecosystem |
| The healthcare ecosystem is the network of stakeholders, processes, and materials necessary to treat an ailment by way of medical intervention on a patient. |
| Stakeholders within the field of healthcare robots can be categorized as 1) primary stakeholders, 2) secondary stakeholders, and 3) tertiary stakeholders. |
| Primary stakeholders use healthcare robotics regularly or even daily and include Direct Robot Users, Clinicians, and Caregivers. |
| Secondary stakeholders are involved in using healthcare robotics but will not directly use them themselves, including Robot Makers, Environmental Service Workers, and Health Administrators. |
| Tertiary stakeholders are those interested in the use and deployment of healthcare robotics in society, although it is unlikely that they will use them directly and include Policy Makers, Insurers and Advocacy Groups. |
| **Chapter 3:** Healthcare Robot Categories |
| Healthcare robots hold the potential to advance social goals in healthcare. |
| Healthcare robots have different embodiments and perform various tasks, including surgical, assistive, and service. |
| There is no universal standard defining the progressive medical robots' autonomy levels. |
| Areas influencing the overall safety of robots requiring urgent attention are those of different embodiments, healthcare robot autonomy levels, user psychological aspects, and diversity and inclusion. |
| Assistive robotics can be divided between socially and physically depending on how they assist a user. |
| Industrial robots working for the healthcare industry may need an added safety layer. Depending on the robot task, healthcare robots may provide more autonomy and independence to the users. |
| *Section 1: Surgical Robots* |
| While research has stated that the more autonomous medical robots become, the less human oversight will be, RAS is not gearing towards humanless surgeries. Even though surgery robots operate increasingly autonomously, humans will still perform many tasks and play an essential role in surgery. |
| Robotic platforms for surgical procedures involve an interplay between the sophisticated automated platform, on the one hand, and the surgeon, along with his/her team, on the other. The outcome of such shared task performance essentially depends on how they can be attuned to one another. |
| While increasingly autonomous robotic assistance levels allow intricate surgical feats to be performed with a smaller chance of human error, in surgeries that are very high volume, human surgeons still outperform robots in terms of weighing experience, making complex surgical judgments and developing contextual understanding. |
| The absence of a clear framework that establishes a common minimum baseline for robotic surgeons creates incongruity in procedural safety. A sound basis for such a framework demands a clear and elaborate overview of the most significant existing categories of surgical robots, their intended purpose, use, and main characteristics. |
| While the benefits RAS provides within the field of MIS appear to be equally distributed among males and females, the research identifies that females generally express more concerns concerning the safety and perception of new technology in surgery than males, who tend to be more unfazed by the notion of robotic surgery. This may be due to their different understanding and conceptualization of RAS. |
| *Section 2: Assistive Robots* |

| | |
|---|---|
| **Socially-assistive robots** | The industry navigates numerous entanglements between the patient's needs, the translation of those needs into concrete assistance, and how social interaction can modulate robots' assistance. The simultaneous fluidity and opaqueness of these entanglements lead to a confusing but fruitful development of devices that flirt the boundary between the categories of medical device, toy, and product, very much demanding a precise definition, |
| | categorization, and analysis to determine the future role and scope of action of the law. |
| | A general division of socially assistive robots into therapy robots and care robots provides a more explicit framework that accentuates the need for manufacturers to state such robots' *medical intended purpose* to avoid misclassification and insufficient safeguards to ensure safe use. |
| | SARs state of the art illustrates the definitional opposition between the robot categories of care and therapy. One-on-one correspondences between the intended purpose and context of use are only visible for the robots falling under the therapy category. |
| | Recognizably gendered robots lead to the projection of gendered stereotypes onto those robots. Still, having robots without recognizable human traits such as gender could further the intrinsic dehumanization of a non-human agent taking over the human role of care. |
| | Sex robots can be a tool to help realize the sexual rights of persons with disabilities. However, it is essential to anticipate the specific physical and sex-related needs that disabled persons might have before designing sex care robots. |
| **Physically assistive robots** | The demand for physical therapy services worldwide has increased in recent years, partly due to aging populations. Assistive technologies and rehabilitation robotics have become popular, especially as they promise to ease the stress on medical and physiotherapy staff and control expenses improving PCN-impaired individuals' lives. |
| | The original and reviewed ISO standard provides a reasonable basis for defining PARs and accounts for the need to have specific requirements for different user categories. However, they fail to consider concrete (often more vulnerable) users, bringing about uncertainties about exactly what falls within the framework's protected scope. |
| | Although PAR categories are strictly separate from one another and other healthcare robot categories, in theory, they appear to be of a gradient nature in practice, often overlapping in their purpose and function. As a result, this may lead to a discrepancy in which regulations govern these devices of the same fundamental nature, i.e., the medical device regulation or the personal care standard. |
| | There currently exists blurriness between robotic nurses as merely physical assistants and the growing desire for multipurpose and multi-functional robots, possibly leading to the existence of a mixed assistance category/trend heading in the direction of mixed assistance. |
| | Policymakers should avoid the scenario where user's safety is defined and enforced in varying ways depending on the producer's end goal. |
| | Robotic engineers develop robotic devices primarily on the premise that the product should fit as many end-users as possible, ultimately serving the majority of users but failing to address the needs of a broader spectrum of users, including those that deviate from the majority of users. |
| | The question of device acceptance is essential when considering the growing ethical implications of the robotization of care. PARs could potentially reinforce the notion that care can be reduced to merely caring for bodies, as well as impacting their dignity and lessening their self-respect. |

| |
|---|
| *Section 3: Healthcare Service Robots* |
| The deployment of HSRs significantly increases overall efficiency in the healthcare sector while simultaneously protecting front-line hospital staff from the risk of disease, eliminating human error, and allowing staff to focus their attention on higher priority tasks. |
| HSRs require lower-cost sensor alternatives to make them more accessible and cost-effective. If this need is met, job restructuring is likely to occur due to the increased HSR employment rates. |
| While HSRs can be significantly beneficial to the healthcare sector, this development, in turn, raises novel occupational safety and health issues. It is also unclear whether existing legal frameworks and guidelines entail sufficient adequate responses to health safety-related issues caused by the deployment of HSRs, which are increasingly autonomous, in uncontrolled environments. In this sense, policies and (additional) safeguards ought to be put in place to ensure that robot manufacturers follow sufficiently adequate legal frameworks as other industries do. |
| The COVID-19 pandemic has allowed better, cheaper, and more accessible HSR. Policymakers are encouraged to take regulatory action sooner or later to address the (adverse) implications of the rapid adoption of these technologies and the compromises on user safety that have been made due to exceptional circumstances. |
| To promote positive attitudes and acceptance towards HSR, public engagement campaigns to introduce HSR to society and train healthcare staff on how to operate them would foster trust and societal engagement. |
| It remains unclear who the actual users of HSRs are and how these service robots 'fit' into existing social systems. While it is thus crucial that HSRs are compatible with existing processes and systems in the hospital context, it is equally important that these devices are compatible with the attitudes and values present in the hospital context. |

## The key take-aways for policymakers

| **On Healthcare Robots \| key take-aways for policymakers** |
|---|
| **Healthcare Robots** |
| Healthcare robots differ from other types of robots and deserve special regulatory attention. |
| Medical robots' embodiment and capabilities differ vastly across surgical, physically/socially assistive, or serviceable contexts and the involved human-robot interaction is also very distinctive, which demands a risk-based and case-by-case analysis. |
| Despite several clear benefits, healthcare robots exert direct control over the world can cause harm in a way that humans cannot necessarily correct or oversee. Therefore, the parameters in which we assess the need for healthcare robots cannot be resource efficiency and increased productivity as with industrial robots, as healthcare involves greater values at stake, including patient safety. |
| Policymakers should consider robots and AI for healthcare purposes to be considered 'high-risk' applications within proposed regulatory frameworks (e.g., AI Act 2021). |
| The industry and their industrial standards are pushing for new in-between categories in the healthcare sector (personal care) outside the medical device categorization. This creates enormous legal uncertainty in ensuring robot and patient safety. |
| The *levels of automation* define the robot's progressive ability to perform particular functions independently. While the Society of Automotive Engineers established automation levels to clarify the progressive development of automotive technology, no universal standards have been defined for progressive autonomy levels for medical robots. |
| Although some of the ethical, legal, and societal concerns with respect to healthcare robots are shared with other types of robots and information technologies, healthcare robots' unique combination of features raises specific issues that deserve attention. |
| *Surgery Robots* |
| While fully autonomous surgical robotic devices have not yet found their way into RAS, they are likely to enter clinical practice, albeit in the distant future, raising myriad legal questions and concerns. |
| Even though surgery robots operate increasingly autonomously, humans will still perform many tasks and play an essential role in determining the robot's course of operation. Understanding the exact role of medical staff in robot-assisted surgeries is essential to understand better who is responsible if something goes wrong. |
| While robot-assisted surgeries have become the first-choice surgical modality for several surgical procedures, research has shown that patient's and medical staff's RAS perceptions often do not accurately reflect reality. Understanding the different perceptions of different users regarding robot-assisted surgeries plays an important role in developing surgical robots. This also applies to surgery-specific developments in RAS. |
| There are currently no standardized training modules for the use of surgical robots. The importance of having a validated training curriculum in place not only springs from the responsibility towards patient safety but also the ensuing issues with credentialing and associated liability. |
| *Assistive Robots* |
| *Socially Assistive Robots* |
| SARs state of the art illustrates the definitional opposition between the robot categories of care and therapy. The very existence of care as a domain of application of healthcare technologies allows for the blurriness in the robots' application, notably revealing an unclear |

| boundary between robots for serviceable contexts and healthcare robots. These blurred contexts of use make it increasingly difficult to define what counts as a medical device according to the European medical device regulation, thereby presenting unforeseen risks to a user. |
|---|
| The target end-users for most SARs' applications have an overfocus on children. Ideally, SARs' embodiments need to adjust to a wide range of individuals who will come in contact with them and cannot be limited to the narrow pool of subjects determined by research. |
| The use of technology in mental health is somewhat scattered in service of the exploration of robotics and AI applications. The majority of novel robot applications, however, are developed within the context of projects or labs and designed for specific research purposes. In reality, very few robots, whether commercial or experimental, are thus actively used by users in their daily lives. At the same time, the limited scope of application also brings with it substantial repurposing. |
| What is notable regarding SARs' embodiment is the tendency towards child-friendly, if not child-like, and toylike, appearances. Most available SARs navigate between anthropomorphic or zoomorphic forms. Excluding their philosophical implications, anthropomorphic robots' byproduct consequence is the potential exacerbation of human biases, notably that of race and gender. Policies should better account for such implications. |
| The full realization of the sexual rights of persons with disabilities requires more research and policies that understand the intersection of people, disability, and sexual rights. Policies could represent a step forward in treating people with disabilities in a dignity-respecting and non-discriminatory fashion concerning their sexual rights, while also accounting for other unexpected consequences. |
| *Physically Assistive Robots* |
| Although each of the categories established within the field of PARs are strictly separate from one another and from other healthcare robot categories, in theory, they appear to be of a gradient nature in practice, often overlapping one another in their purpose and function. The establishment of fixed categories will likely prevent such complex robots from being properly assessed and addressed by policymakers. |
| There is a broad range of PARs currently available to the public and used within an academic context. However, the introduction of robots for health-care and therapeutic purposes that interact directly with patients, older adults, and children has proved to be complicated. Within the field of PARs, generally, there is an overfocus on adult populations, while a particularly salient gap in access is found in children. To ensure that PARs are representative of all possible users, this overfocus will have to be addressed. |
| PARs are generally characterized by the variety of user interfaces and control systems that they incorporate and user perspectives that they generate. These allow a person with a disability to interact within society and their surroundings and care for themselves more independently. While these machines become more capable, however, they often also become more complex. As control systems (continue to) become increasingly sophisticated and allow for larger "bandwidths" of information to be transferred from human to machine, increasingly sophisticated devices will be developed. In all cases, the control software must generate motion directives that realize the required action while considering safety, comfort, and efficiency. |
| The development of robotic assistance is often orchestrated with the needs of caregivers instead of care receivers in mind, with user autonomy being pitched as a problem to be solved rather than a faculty to be developed. The replacement of human assistance by robotic assistance to realize ADL. The question of device acceptance is essential when considering the growing ethical implications of the robotization of care. |
| Sex robots could represent a valuable tool to help realize the sexual rights of persons with disabilities, although more research is needed in this respect. |
| *Healthcare Service Robots* |
| Although it is becoming increasingly clear that the deployment of HSRs significantly increases overall efficiency in the healthcare sector while simultaneously protecting front-line hospital staff from the risk of disease, eliminating human error, and allowing staff to focus their attention on higher priority tasks, a move towards lower-cost sensor alternatives should be considered to make these robots more accessible. |
| While HSRs can be significantly beneficial to the healthcare sector, they, in turn, raise novel occupational safety and health issues. It is unclear whether existing legal frameworks and guidelines entail sufficient adequate responses to health safety-related issues caused by the deployment of HSRs, which are increasingly autonomous, in uncontrolled environments. In this sense, policies and (additional) safeguards ought to be put in place to ensure that robot manufacturers follow sufficiently adequate legal frameworks as other industries do. |
| The deployment of robots during the COVID-19 pandemic is happening quicker than expected. The COVID-19 pandemic has allowed the deployment of these technologies to become better, cheaper, and more accessible. In this respect, policymakers will be forced to catch up and take regulatory action sooner or later to address the (adverse) implications of the rapid adoption of these technologies and the compromises on user safety that have been made due to exceptional circumstances. |
| A significant challenge in the field of HSRs is their acceptance, especially with regard to job maintenance. To promote positive attitudes and acceptance towards HSR, public engagement campaigns to introduce HSR to society and train healthcare staff on how to operate them would foster trust and societal engagement. |
| There is a lack of clarity regarding the full potential of service robots in healthcare. Moreover, it remains unclear who the actual users of HSRs are and how these service robots 'fit' into the existing social system. While it is thus crucial that HSRs are compatible with existing processes and systems in the hospital context, it is equally important that these devices are compatible with the attitudes and values present in the hospital context. |

**A list of policy recommendations**

Our findings point to the following policy recommendations:

1. Healthcare robots differ from other types of robots and deserve special regulatory attention.

2. Legal frameworks should acknowledge the high-risk nature of robotic systems for healthcare purposes.

3. Clarity on the medical device categorization for healthcare robot types, including surgical, physically and socially assistive, and healthcare service robots is needed at risk that they will be repurposed without the appropriate safeguards.

4. Universal standards for progressive autonomy levels for healthcare robots should be defined to increase legal certainty with respect to responsibility allocation in highly complex human-robot interaction contexts, also in the medical field.

5. The parameters in which the need for healthcare robots is assessed cannot be resource efficiency and increased productivity as with industrial robots only, other aspects as the implications for society need to be considered.

6. To address safety comprehensively, healthcare robots demand a broader understanding of safety, extending beyond physical interaction, but covering aspects such as cybersecurity, temporal aspects, societal dimensions and mental health.

7. Embodied healthcare robots can also exacerbate existing biases against certain groups and, therefore, their design, implementation, and use should account for diversity and inclusion.

# About the authors

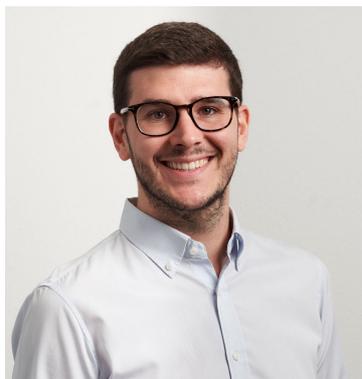

**Dr. Eduard Fosch-Villaronga**

Dr. Fosch-Villaronga is an Assistant Professor at the eLaw Center for Law and Digital Technologies at Leiden University (NL) where he investigates legal and regulatory aspects of robot and AI technologies, with a special focus on healthcare. Eduard recently published the book 'Robots, Healthcare, and the Law. Regulating Automation in Personal Care' with Routledge and is interested in human-robot interaction, responsible innovation, and the future of law.

Eduard is the PI of LIAISON, an FSTP from the H2020 COVR project that aims to link robot development and policymaking to reduce the complexity in robot legal compliance. He is also the PI of PROPELLING, an FSTP from the H2020 Eurobench project, a project using robot testing zones to support evidence-based robot policies. Eduard is the co-leader of the Ethical, Legal, and Societal Aspects Working Group at the H2020 Cost Action 16116 on Wearable Robots and participates actively at the Social Responsibility Working Group at the H2020 Cost Action 19121 GoodBrother. In 2020, Eduard served the European Commission in the Sub-Group on Artificial Intelligence (AI), connected products and other new challenges in product safety to the Consumer Safety Network (CSN) to revise the General Product Safety directive.

Previously, he worked as a Marie Skłodowska-Curie Postdoctoral Researcher under the LEaDing Fellows at eLaw (Jan 2019-Dec 2020). He also was a postdoc at the Microsoft Cloud Computing Research Center at Queen Mary University of London (the UK, 2018) investigating the legal implications of cloud robotics; and at the University of Twente (NL, 2017) as a postdoc, exploring iterative regulatory modes for robot governance. Eduard Fosch-Villaronga holds an Erasmus Mundus Joint Doctorate (EMJD) in Law, Science, and Technology coordinated by the University of Bologna (IT, 2017), an LL.M. from University of Toulouse (FR, 2012), an M.A. from the Autonomous University of Madrid (ES), and an LL.B. from the Autonomous University of Barcelona (CAT, 2011). Eduard is also a qualified lawyer in Spain and his publications are available online.

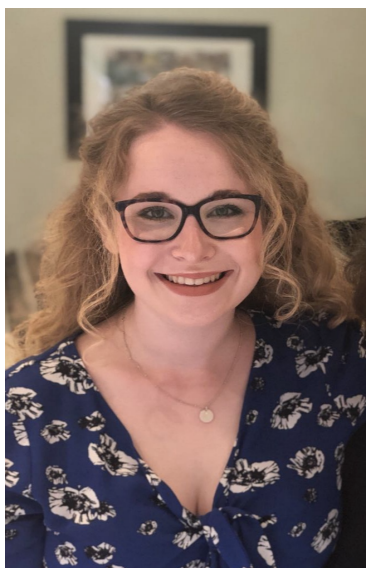

**Ms. Hadassah G. Drukarch**

Hadassah Drukarch is a Research Assistant at the eLaw Center for Law and Digital Technologies at Leiden University (NL), and the founder of The Law of Tech, an online platform aimed at preparing the legal world of tomorrow by educating about the interaction between law and new technologies. She is currently working on Liaison, an FSTP from the H2020 COVR project that aims to link robot development and policymaking to reduce the complexity in robot legal compliance under the EU's H2020 Research and Innovation Program Grant Agreement No 779966.

Hadassah's work navigates between Artificial Intelligence, Robots, Law, Business, and Governance, Diversity and Human

Rights. She wrote a dissertation on 'Robots, Directors, and Liability: Understanding corporate (mis)conduct in the context of robot-directors,' to conduct further research within the field. In 2019, she was an intern at the legal department of the LegalTech start-up Ligo, a legal-tech startup in the field of corporate law and contract life management (Amsterdam, NL). Hadassah was a research trainee at the eLaw Center for Law and Digital Technologies in 2020, working on Healthcare Robotics, including Robotic Assisted Surgery (RAS) regulatory and liability aspects. Hadassah holds an LL.B. in International Business Law and an Honours College Law Program certificate from Leiden University. In Sept 2021, Hadassah will take part in the Advanced LL.M. in Law and Digital Technologies at Leiden University.